\documentclass{llncs}

\usepackage{eccv}

\usepackage{eccvabbrv}

\usepackage{graphicx}
\usepackage{booktabs}
\usepackage{microtype} % added
\usepackage{subcaption} % added
\usepackage{amsfonts} % added
\usepackage[colorinlistoftodos]{todonotes} % added

\usepackage[accsupp]{axessibility}  % Improves PDF readability for those with disabilities.

\usepackage{hyperref}

\usepackage{orcidlink}
\usepackage{enumitem} % added

\usepackage{mathtools}
\usepackage{multirow}
\usepackage{pifont}
\newcommand{\cmark}{\ding{51}}

\graphicspath{{figures/}}
\usepackage{tikz}
\usetikzlibrary{positioning, arrows.meta, shapes.geometric, fit, backgrounds, calc}

\begin{document}

\title{TrianguLang: Geometry-Aware Semantic Consensus for Pose-Free 3D Localization} 

% TODO REVIEW: If the paper title is too long for the running head, you can set an abbreviated paper title here. If not, comment out.
\titlerunning{TrianguLang}

\author{Bryce Grant \and
Aryeh Rothenberg \and Atri Banerjee \and Peng Wang}

\authorrunning{B. Grant et al.}

\institute{Case Western Reserve University, Cleveland, OH, USA\\
\email{\{bag100, pxw206\}@case.edu}}

\maketitle

\begin{abstract}
 Localizing objects and parts from natural language in 3D space is essential for robotics, AR, and embodied AI, yet existing methods face a trade-off between the accuracy and geometric consistency of per-scene optimization and the efficiency of feed-forward inference. We present TrianguLang, a feed-forward framework for 3D localization that requires no camera calibration at inference. Unlike prior methods that treat views independently, we introduce Geometry-Aware Semantic Attention (GASA), which utilizes predicted geometry to gate cross-view feature correspondence, suppressing semantically-plausible but geometrically-inconsistent matches without requiring ground-truth poses. Validated on five benchmarks including ScanNet++ and uCO3D, TrianguLang achieves state-of-the-art feed-forward text-guided segmentation and localization, reducing user effort from $O(N)$ clicks to a single text query. The model processes each frame at 1008x1008 resolution in $\sim$58ms ($\sim$17 FPS) without optimization, enabling practical deployment for interactive robotics and AR applications. Code and checkpoints are available at \url{https://cwru-aism.github.io/triangulang/}.

% We present TrianguLang, a feed-forward framework for part-level 3D localization that requires no camera calibration at inference. Our key insight is that predicted geometry from foundation models can serve as an "arbiter" for semantic attention, suppressing false correspondences that arise when views disagree. We introduce Geometry-Aware Semantic Attention (GASA), a novel attention mechanism that combines semantic similarity with geometric proximity, attending only to features that are both semantically similar and geometrically close in 3D space. Combined with world-space positional encoding and spatial language parsing, GASA enables zero-shot transfer to novel scenes without camera calibration. Across five benchmarks: ScanNet++, uCO3D, NVOS, SPIn-NeRF, and PartImageNet; Triangulang achieves state-of-the-art performance on text-guided multi-view segmentation while reducing user effort from $O(N)$ clicks to a single text query. TrianguLang operates at $\sim$10 FPS with no per-scene optimization, enabling practical deployment for interactive robotics and AR applications.
\end{abstract}

\section{Introduction}
\label{sec:introduction}
Promptable segmentation has developed recently, enabling complex scene understanding through simple visual prompts like text and bounding boxes. SAM \cite{sam} established the foundation by demonstrating promptable segmentation across diverse applications, with subsequent versions introducing further features such as video segmentation \cite{sam2} and text-based querying \cite{sam3}. Nevertheless, despite impressive performance, these models lack 3D awareness, making them prone to object flickering and occlusions.

Incorporating geometric awareness into promptable segmentation presents fundamental challenges. Existing geometric representations (e.g. point clouds, voxels, 3DGS \cite{3dgs}, NeRFs \cite{nerf}) enable lifting 2D information into 3D, but require lengthy per-scene optimization and prioritize fidelity over semantic understanding. Additionally, current promptable segmentation models cannot estimate object position in 3D space, while pose estimation methods \cite{conceptpose,foundationpose} require preexisting masks, bounding boxes, or cropped images as input.
% -challenges: they face long per-scene optimization (i.e. setup time).

%When writing about another work, capture its importance from the perspective of your own study and contributions
Neural rendering is another recent development that has transformed 3D scene reconstruction. NeRF~\cite{nerf} introduced a novel view synthesis technique requiring only sparse input images with known camera poses. 3D Gaussian Splatting~\cite{3dgs} followed, replacing dense volume representations with lighter anisotropic Gaussians and achieving real-time rendering at a fraction of the compute. While impressive, 3DGS retained reliance on calibrated camera poses and did not address the prospect of downstream tasks like object detection and semantic scene understanding. Thus, as reliable and context-free backbones for 3D localization and reconstruction, NeRF and 3DGS present promising foundations for advancing 3D object awareness, segmentation, and scene understanding.
%like NeRF, 3DGS relies on posed image inputs and is not designed for downstream tasks such as object detection or semantic scene awareness).  3DGS does nothing to alleviate NeRF's reliance on multiple image inputs, nor, more importantly, its complete lack of object recognition or intrinsic scene comprehension.

Fusion models have emerged as successful extensions, distilling reconstruction techniques with features from foundation models like CLIP \cite{clip} and SAM \cite{sam}. One approach utilizes natural language grounding to embed high-dimensional language features into 3D space, enabling open-vocabulary querying of reconstructed scenes \cite{lerf, langsplatv2, openscene3d, fmgs}. Similarly, identity-based approaches assign semantic attributes to individual scene primitives (e.g., Gaussians), allowing promptable 3D instance segmentation~\cite{gaussiangrouping, SAGA, refersplat}. Last, spatial reasoning approaches like SpatialRGPT~\cite{spatialrgpt} and 3D-LLM~\cite{3dllm} inject geometric information into large language models for complex spatial queries.
% Another approach relates parts of a scene by assigning identities to the individual building blocks (e.g. the Gaussians), allowing for precise, promptable, 3D instance segmentation and object recognition \cite{gaussiangrouping, SAGA, refersplat}. *Add in VLM approaches such as LEO. Similarly, SpatialRGPT and 3D-LLM inject geometric information into the LLM...

%storing semantic attributes alongside original parameters
%integrating CLIP and DINO into the GS (Langsplat)
%promptable GS using SAM3 (refersplat)

Visual geometry models (VGMs) \cite{da3, pi3, dust3r, VGGT} take a step beyond traditional NeRFs and 3DGS, demonstrating that scene reconstructions can be produced from unposed RGB images in an end-to-end process.
%For 3D localization, monocular depth estimation provides metric depth from a single image \cite{Marigold, da3}.
Nevertheless, these models, too, lack intrinsic scene awareness, presenting the opportunity for a new framework capable of handling both semantic understanding and 3D localization without per-scene optimization.
% How can we segue into presenting our model?
%VGGT, Pi3, DA3 can all do better reconstructions from single imagw

% You will see papers with preliminaries that provide the mathematical foundation for a topic that is discussed in methodology. That being said, DO NOT put overly complex explanations and/or math in the introduction

% \todo[inline]{Maybe add benefits associated with pointmaps derived due to bijective mapping of pixels to 3d points.}

Bridging these two fields, we propose \textbf{TrianguLang}, a language-guided multi-view segmentation and 3D localization framework. We blend semantic knowledge from SAM3~\cite{sam3} with geometric priors from DA3~\cite{da3} through our novel \textbf{Geometry-Aware Semantic Attention (GASA)} decoder, which uses world-space positional encoding and geometric attention biasing to achieve cross-view consistency without explicit correspondence supervision.

% We evaluate TrianguLang on 6 diverse benchmarks:
% \begin{itemize}[nosep,leftmargin=*]
%     \item \textbf{ScanNet++}~\cite{scannet++}: Indoor scenes with 450+ rooms, instance-level annotations
%     \item \textbf{UCo3D}~\cite{UCo3D}: Object-centric 360° captures, 170K videos across 1000 categories
%     \item \textbf{MVImgNet}~\cite{MVImgNet2}: 520k multi-view images of objects
%     \item \textbf{NVOS}~\cite{NVOS}: Multi-view video object segmentation
%     \item \textbf{SpinNeRF}~\cite{spinnerf}: Object-centric captures for part-level evaluation
%     \item \textbf{PartNeXt}~\cite{partnext}: 350K part annotations across 23K models for part-level grounding
% \end{itemize}

\noindent Our contributions include:
\begin{enumerate}[nosep,leftmargin=*]
    \item \textbf{GASA (Geometry-Aware Semantic Attention)}: A novel attention mechanism that combines semantic similarity with geometric constraints from monocular depth, enabling pose-free cross-view consistency without explicit correspondence supervision.

    % \item \textbf{Sheaf-Theoretic Consistency}: A formalization of multi-view semantic agreement as finding a global section of a cellular sheaf, providing a principled regularizer that encourages topological coherence across views.

    \item \textbf{Pose-Free 3D Localization}: Camera-relative object localization via depth unprojection, providing metric 3D coordinates (e.g., ``1.2m ahead, 0.3m left'') without SLAM or camera pose estimation.

    \item \textbf{LLM-Free Spatial Language}: Support for spatial qualifiers and relational queries via direct geometric computation without requiring LLM inference, enabling real-time spatial grounding.
\end{enumerate}

\section{Related Work}
\label{sec:related}
% \todo[inline]{ATRI \& ARYEH: Look into oracle IoU/mACC for these segmentation models. The problem with promptable segmentation is if you provide a prompt like apple or click an apple and there are multiple apples in the scene, if you take the top k masks produced by your model, then it will segment all apples or it might segment the wrong apple (the one you did not intend). So many segmentation models rely on Oracle IoU which means at run time, they pick the masks that optimize the IoU (compare to GT) If you do find papers reporting both, feel free to put a table in discussing. }

\subsection{Optimization-Based 3D Grounding}
Traditional methods for grounding language in 3D, such as ScanRefer \cite{scanrefer} and OpenScene \cite{openscene3d} operate on pre-scanned 3D point clouds and posed images as input. While effective, they are limited by the requirement of a pre-existing, high-quality 3D map.

More recent neural rendering approaches achieve dense 3D grounding by distilling semantic features into implicit representations. LERF~\cite{lerf} embeds multi-scale CLIP features into NeRF via volume rendering, enabling open-vocabulary 3D queries. LangSplat~\cite{langsplat} extends this to 3D Gaussian Splatting by using SAM to segment training images, extracting hierarchical CLIP features, and attaching compressed representations to each Gaussian. FMGS~\cite{fmgs} uses multi-resolution hash encoding to represent CLIP and DINO features more efficiently. LEGS~\cite{yu2024legs} and SemanticSplat~\cite{semanticsplat} similarly augment Gaussians with semantic attributes distilled from foundation models. While SemanticSplat employs feed-forward feature prediction, it still requires a pre-computed 3DGS reconstruction from calibrated images. These methods share fundamental limitations: they require calibrated camera poses and per-scene reconstruction or optimization (10 to 45 minutes). TrianguLang operates in a feed-forward manner on uncalibrated images, processing new scenes without any optimization.

Any6D \cite{any6d} estimates the 6D pose of novel objects in a feed-forward manner but operates at object-level granularity and lacks language supervision. Recent work such as Segment Any 3D Gaussians \cite{SAGA} and Click-Gaussian \cite{clickgauss} extend 3D segmentation to Gaussian Splatting but require clicks or pre-computed masks rather than end-to-end language grounding.

% \todo[inline]{The use of CLIP for text based segmentation/localization. And the downsides of using CLIP in depth estimation}

\subsection{Open-Vocabulary 3D Segmentation}
State-of-the-art open-vocabulary methods like OpenMask3D~\cite{openmask3d} and Open3DIS~\cite{open3dis} adopt a ``lift-and-project'' strategy: they segment 2D images and aggregate features into a pre-built 3D point cloud. OmniSeg3D~\cite{omniseg3d} uses hierarchical contrastive learning for omniversal segmentation, but still relies on dense 3D reconstruction. These methods cannot function without a pre-existing 3D map of the environment. ReferSplat~\cite{refersplat} introduces Referring 3D Gaussian Splatting Segmentation (R3DGS), which segments objects based on natural language descriptions containing spatial relations and attributes. ReferSplat demonstrates that 3DGS can handle complex referring expressions with spatial language. However, it still requires per-scene optimization ($\sim$58 minutes training) and calibrated poses for reconstruction. TrianguLang achieves similar spatial language capability in a feed-forward, pose-free setting via explicit geometric computation on depth-derived 3D positions without requiring reconstruction.

% Each Gaussian receives a learned ``referring feature'' vector; segmentation masks are obtained by computing similarity between Gaussian features and BERT-encoded text embeddings. A Position-aware Cross-Modal Interaction module integrates 3D position information into cross-attention, enabling spatial reasoning (e.g., ``the cup near the window''). Gaussian-Text Contrastive Learning further improves discrimination between similar expressions.

\subsection{Multi-View Segmentation}
SAM2 and SAM3 \cite{sam2,sam3} propagate masks through video but lack 3D geometric guarantees, drifting when views change significantly. MV-SAM  \cite{mvsam} leverages pointmaps for multi-view segmentation but relies on click prompts rather than language. MV-SAM treats views as independent samples, admitting no explicit enforcement of geometric consistency. GTA \cite{GTA} injects camera geometry into attention via epipolar constraints but requires \textit{known} camera poses. Multi-View Foundation Models \cite{mvfoundation} adapt 2D backbones into multi-view consistent variants via cross-view attention and Pl\"ucker pose embeddings. However, like GTA, they require known camera poses and focus on feature consistency rather than language-prompted localization, outputting per-pixel embeddings rather than 3D coordinates. Emerging approaches like VGGT \cite{VGGT} explore visual-geometric grounding with transformers but focus on generation rather than metric localization. We adapt this insight to the map-free regime by driving attention with estimated metric depth, enabling geometry-aware consistency without requiring a pre-built dense map.

% but requires click prompts, admits no explicit 3D consistency enforcement and outputs masks, not 3D coordinates.

\subsection{Spatial Reasoning in Vision-Language}
Interpreting spatial relations (``nearest,'' ``left of,'' ``behind'') in natural language is challenging. Recent approaches delegate this to large models: SpatialVLM~\cite{spatialvlm} fine-tunes a 7B-parameter VLM on 2M spatial QA pairs; 3D-LLM~\cite{3dllm} injects 3D point cloud features into LLaMA; LEO~\cite{leo} uses an embodied agent with chain-of-thought prompting; SpatialRGPT~\cite{spatialrgpt} conditions a VLM on depth; LLM-Grounder~\cite{llm-grounder} uses an LLM agent to iteratively refine 3D bounding boxes. SAM3's Agent mode~\cite{sam3} combines with an MLLM (Qwen2.5-VL) for complex spatial queries.

These approaches achieve impressive reasoning but incur 1 to 10+ seconds latency per query. Concurrently, mechanistic interpretability work~\cite{spatialids2026} reveals that VLMs encode spatial locations by linearly binding ``Spatial IDs'' to object token activations, but the work finds that depth and height share similar spatial representations, suggesting fundamental limitations in 3D reasoning.

TrianguLang takes a complementary approach: rather than relying on learned spatial representations that suffer from this depth collapse, we parse a constrained vocabulary of spatial qualifiers (nearest, leftmost, behind, etc.) with regex and resolve them via direct geometric computation on depth-derived 3D centroids. This enables real-time spatial grounding ($\sim$100ms) with true 3D awareness without relying on LLMs.

% \begin{itemize}
%     \item \textbf{Gaussian Grouping \cite{gaussiangrouping}}: Assignment of identity encodings to individual Gaussians to enable element grouping and object-level scene understanding. Class labels predicted from 2D input by SAM. Grouping allows for scene editing: "3D object removal, inpainting, colorization, style transfer and scene recomposition."
%     \item \textbf{SemanticSplat \cite{semanticsplat}}: Augments individual Gaussians with semantic attributes during rendering. Knowledge distillation from SAM and CLIP-LSeg allows for promptable, open-vocabulary segmentation. % not sure how we're structuring RR anymore. This could go in O-V Seg.
%     \item \textbf{GSNeRF \cite{gsnerf}}: Derives visual features alongside depth information to build semantic segmentation maps into NeRFs.
%     \item \textbf{FMGS \cite{fmgs}}: Semantic embedding of CLIP and DINO feature maps in Gaussian Splats. Uses Multi Resolution Hash encodings to increase efficieny by avoiding per-item labeling (I think).
%     \item \textbf{ReferSplat \cite{refersplat}}: Assigns each Gaussian a referring feature vector and then determines the similarity with textual queries (natural language expressions) to develop a 2D segmentation mask. Uses contrastive learning on the Gaussians and the text expression to ensure the query leads to the correct object.
% \end{itemize}

% Semantic Gaussian Splatting subsection removed for space (content covered in Optimization-Based section above)

\section{Method}
\label{sec:method}
\subsection{Problem Formulation}
Given $N$ RGB images $\{\mathbf{I}_i\}_{i=1}^N$ of a scene and a text query $\mathcal{T}$ (e.g., ``the red mug''), our goal is to predict:
\begin{enumerate}[nosep,leftmargin=*]
    \item Per-view binary masks $\{\mathbf{M}_i \in \{0,1\}^{H \times W}\}_{i=1}^N$
    \item A world-frame 3D centroid $\mathbf{c} \in \mathbb{R}^3$ localizing the queried object
\end{enumerate}

We do \textbf{not} require ground-truth camera intrinsics or extrinsics. Prior multi-view methods assume calibrated cameras~\cite{mvsam}, while TrianguLang uses \emph{predicted} geometry from DA3-NESTED~\cite{da3}, which jointly estimates metric depth, intrinsics, and extrinsics from the input images alone. This enables deployment on uncalibrated camera systems and multi-camera setups without requiring SLAM or structure-from-motion (SfM) preprocessing.

Unlike prior work that requires per-view text prompts or click supervision, TrianguLang operates in a \emph{single-prompt} setting: the language query is provided once and propagated across all views through geometric reasoning, reducing user effort from $O(N)$ clicks to $O(1)$ text input. While recent approaches to spatial language understanding (e.g., SpatialVLM~\cite{spatialvlm}, 3D-LLM~\cite{3dllm}, LEO~\cite{leo}) require large language models or multi-step agent pipelines to interpret queries like `nearest chair' or `mug to the left of the keyboard,' TrianguLang resolves spatial relations \textit{directly} via depth-derived 3D geometry, eliminating LLM inference latency and enabling real-time spatial grounding.

% Rather than trusting semantic similarity alone, we use geometry as a sanity check, ensuring that two pixels should only correspond if they are both semantically similar AND geometrically proximate.

\subsection{Architecture Overview}
\begin{figure}[htbp]
\centering
\includegraphics[width=\columnwidth]{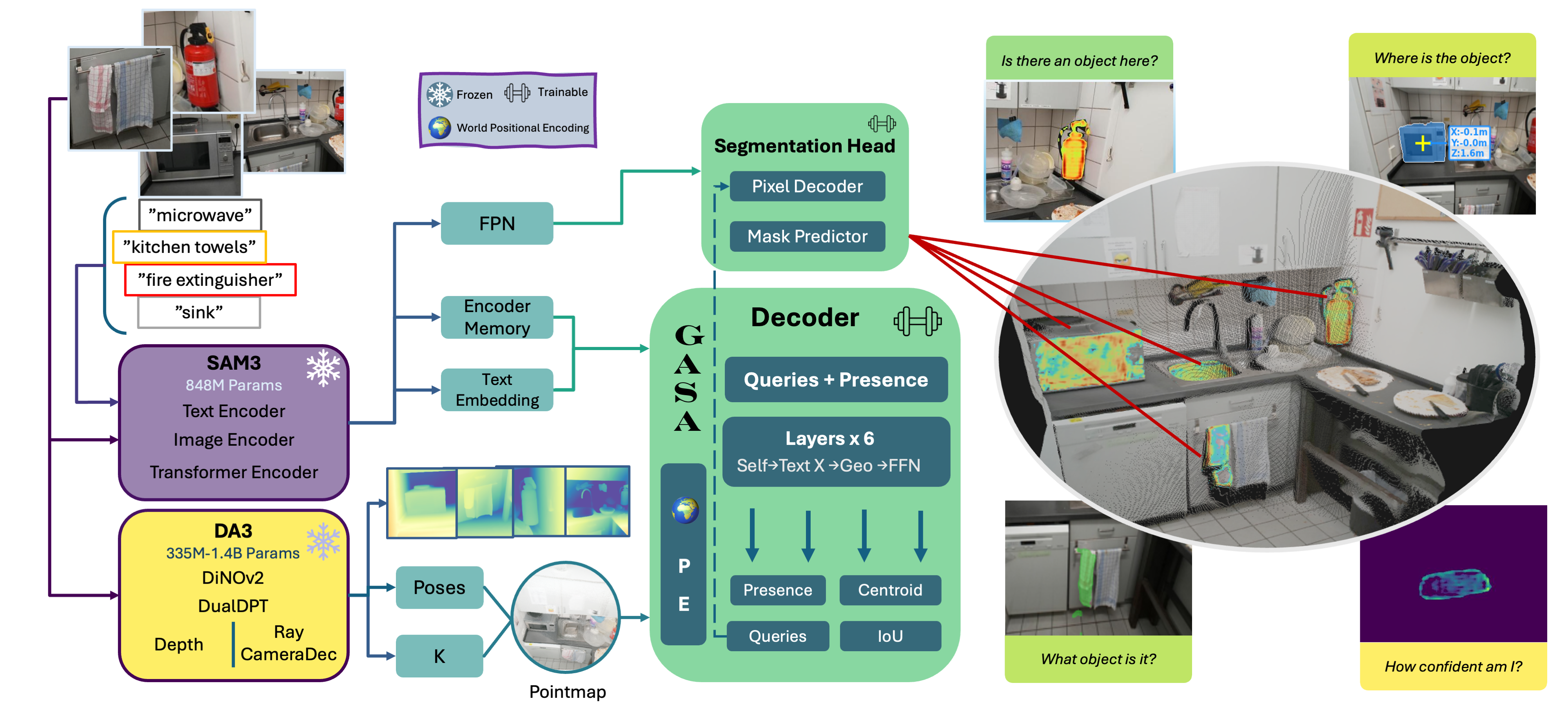}
\caption{Overview of the TrianguLang architecture}
\label{fig:arch}
\end{figure}

Figure~\ref{fig:arch} illustrates the TrianguLang architecture.

% Images are processed through two frozen foundation models: SAM3 for semantic features and DA3 for depth and pose estimation. The resulting features and pointmaps are combined through our GASA encoder/decoder, followed by a symmetric centroid head that outputs the final 3D coordinates. %

TrianguLang comprises three components (Figure~\ref{fig:arch}):
\begin{enumerate}[nosep,leftmargin=*]
    \item \textbf{SAM3 backbone} (frozen, 848M): Text-conditioned semantic features
    \item \textbf{DA3-NESTED depth model} (frozen, 1.4B): State-of-the-art metric depth and pose estimation~\cite{da3}. While other visual geometry models (VGMs) such as DUST3R~\cite{dust3r} and $\pi^3$~\cite{pi3} provide pointmaps and relative poses, DA3 additionally outputs \emph{metric} depth calibrated in meters, which is essential for accurate 3D localization and the distance-based GASA kernel.
    \item \textbf{GASA decoder} (trained, 13.7M): Geometry-aware cross-view fusion
%     \item \textbf{Centroid head} (trained): Direct 3D localization
% \todo[inline]{Centroid head or Segmentation head?}
\end{enumerate}

% \subsubsection{Frozen Backbones.}
% We leverage two pretrained foundation models:
% \begin{itemize}[nosep,leftmargin=*]
%     \item \textbf{SAM3}~\cite{sam3}: A grounded segmentation model providing text-conditioned encoder features $\mathbf{F} \in \mathbb{R}^{B \times L \times D}$ where $L = H' \times W'$ and $D = 256$.
%     \item \textbf{DA3}~\cite{da3}: Depth Anything V3 with two deployment modes:
%     \begin{itemize}[nosep,leftmargin=*]
%         \item \textit{DA3NESTED-GIANT-LARGE} (1.4B params): Jointly processes multiple views with cross-view attention, estimating metric depth $\mathbf{D}_i \in \mathbb{R}^{H \times W}$, camera intrinsics $\mathbf{K}_i \in \mathbb{R}^{3 \times 3}$, and extrinsics $\mathbf{T}_i \in SE(3)$ via ray-based pose estimation.
%         \item \textit{DA3METRIC-LARGE} (0.35B params): Monocular metric depth for real-time inference when camera-relative localization suffices.
%     \end{itemize}
% \end{itemize}

\subsection{World-Space Positional Encoding}

Standard 2D positional encodings fail to capture cross-view relationships: the same 3D point projects to different pixel coordinates across views. We address this with \textbf{world-space positional encoding} that assigns identical embeddings to the same 3D location regardless of viewpoint.

For each pixel $(u, v)$ in view $i$, we compute its 3D coordinate using DA3-estimated depth $\mathbf{D}_i$ and camera parameters:
\begin{equation}
\mathbf{P}_i(u,v) = \mathbf{T}_i \cdot \left( \mathbf{D}_i(u,v) \cdot \mathbf{K}_i^{-1} \begin{bmatrix} u \\ v \\ 1 \end{bmatrix} \right)
\label{eq:unproject}
\end{equation}
where $\mathbf{K}_i \in \mathbb{R}^{3\times3}$ denotes camera intrinsics and $\mathbf{T}_i \in SE(3)$ denotes the camera extrinsic transformation. DA3-NESTED-GIANT-LARGE jointly estimates both $\mathbf{K}_i$ and $\mathbf{T}_i$ from image features via a learned camera decoder (predicting field-of-view and relative pose), placing all views in a shared world frame \emph{without ground-truth calibration}. This is used at both training and inference time. Following NeRF~\cite{nerf}, we encode 3D coordinates with sinusoidal positional encoding projected via an MLP to $D=256$.

\subsection{Geometry-Aware Semantic Attention (GASA)}
\label{sec:gasa}

Standard cross-attention matches features based purely on semantic similarity, leading to false correspondences between visually similar but geometrically distant regions (e.g., two identical mugs). GASA introduces an explicit \textbf{geometric veto} mechanism.

\subsubsection{Attention with Geometric Bias.}
GASA augments self-attention among encoder feature tokens with a geometric bias derived from their 3D positions (Eq.~\ref{eq:unproject}). Each encoder token has a known 3D position from depth unprojection; for tokens at positions $\mathbf{P}_Q$ and $\mathbf{P}_K$:
\begin{equation}
\small
\text{Attn}(\mathbf{Q}, \mathbf{K}, \mathbf{V}) = \text{softmax}\left( \frac{\mathbf{Q}\mathbf{K}^\top}{\sqrt{d}} + \beta \cdot \phi(\|\mathbf{P}_Q - \mathbf{P}_K\|_2) \right) \mathbf{V}
\label{eq:gasa}
\end{equation}
where $\beta$ is a learnable scalar and $\phi$ is a learned distance kernel. The distance $\|\mathbf{P}_Q - \mathbf{P}_K\|_2$ is computed in meters from DA3-estimated pointmaps. The kernel outputs strongly negative values for large distances. Learnable mask queries subsequently cross-attend to these GASA-refined encoder features.

\paragraph{Distance kernel $\phi$.} We parameterize $\phi$ as a 2-layer MLP with hidden dimension 32, initialized to approximate $-\log(1 + d)$. This learned kernel can adapt its distance sensitivity during training, unlike fixed-form alternatives such as RBF kernels $\phi_{\text{rbf}}(d) = -d^2 / 2\sigma^2$ or linear decay $\phi_{\text{lin}}(d) = -d$. We ablate these choices in Table~\ref{tab:ablation}; the learned kernel consistently outperforms fixed alternatives, suggesting that the optimal distance-to-bias mapping is task-dependent and benefits from end-to-end learning.

% \begin{figure}[t]
% \centering
% \fbox{\parbox{0.95\columnwidth}{\centering
% \vspace{1.5cm}
% \textbf{[Figure: GASA attention visualization]}\\[2mm]
% \small (a) Standard attention: both mugs attend\\
% (b) GASA: distant mug suppressed by $\phi(d) \ll 0$
% \vspace{1.5cm}
% }}
% \caption{GASA geometric veto. Standard attention attends to both semantically similar mugs. GASA's geometric bias suppresses the distant mug, focusing on the geometrically consistent target.}
% \label{fig:gasa}
% \end{figure}

\subsubsection{Decoder Architecture}
The GASA decoder consists of 6 transformer layers, each containing: (1) self-attention among queries, (2) text cross-attention (per-layer, following SAM3), (3) GASA cross-attention with geometric bias, and (4) feed-forward network. See Appendix~\ref{app:architecture} for details.

\begin{figure}[ht]
\begin{center}
\includegraphics[width=\columnwidth]{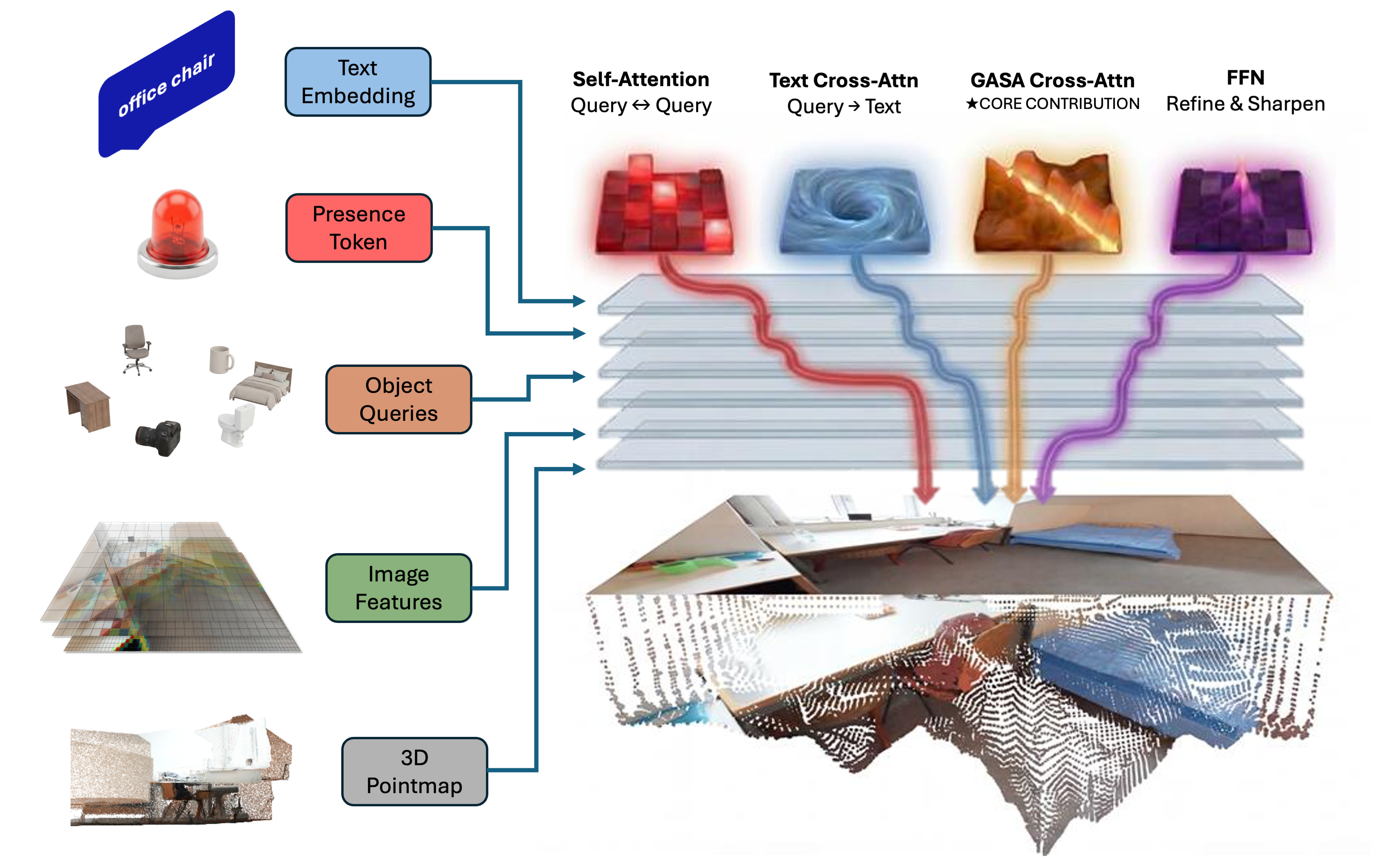}
\end{center}
\caption{Overview of the GASA decoder. }
\label{fig:decoder}
\end{figure}

\subsection{3D Localization}
\label{sec:centroid}

Beyond 2D segmentation, TrianguLang directly predicts 3D object centroids via mask-weighted depth unprojection. For each view $i$:
\begin{equation}
\mathbf{c}_i = \frac{\sum_{u,v} \hat{M}_i(u,v) \cdot \mathbf{P}_i(u,v)}{\sum_{u,v} \hat{M}_i(u,v) + \epsilon}
\label{eq:centroid}
\end{equation}
where $\hat{M}_i \in [0,1]^{H \times W}$ is the predicted mask probability (after sigmoid) for view $i$ and $\mathbf{P}_i$ are DA3-estimated 3D coordinates (Eq.~\ref{eq:unproject}). The final centroid is selected from the view with the highest mask confidence score. The localization pipeline operates entirely on DA3-NESTED estimates, producing 3D coordinates in the estimated world frame.

\subsection{Training Objective}

The complete loss combines segmentation, ranking, and localization terms:
\begin{equation}
\mathcal{L} = \mathcal{L}_{\text{seg}} + \mathcal{L}_{\text{rank}} + \mathcal{L}_{\text{loc}}
\label{eq:total_loss}
\end{equation}

\begin{itemize}[nosep,leftmargin=*]
    \item $\mathcal{L}_{\text{seg}}$: Focal + Dice loss for mask prediction
    \item $\mathcal{L}_{\text{rank}}$: Align loss~\cite{align2024} for confidence-calibrated mask scoring plus contrastive ranking loss for correct mask ordering
    \item $\mathcal{L}_{\text{loc}}$: Smooth L1 centroid regression and binary cross-entropy presence loss for object existence prediction
\end{itemize}

See Appendix~\ref{app:loss} for detailed loss formulations and hyperparameters.

\subsection{Spatial Language Understanding}
\label{sec:spatial}

TrianguLang supports spatial qualifiers (``nearest chair'') and relational queries (``mug left of keyboard'') via geometric computation on depth-derived 3D positions, \textbf{without LLM inference}. The pipeline parses spatial keywords, generates mask candidates via SAM3, computes 3D centroids via depth unprojection, and selects the candidate satisfying the spatial constraint (e.g., $\arg\min_i d_i$ for ``nearest''). During training, we augment labels with spatial qualifiers ($p{=}0.3$) for multi-instance scenes. This enables $\sim$100ms spatial grounding vs.\ 1 to 10+ seconds for LLM-based methods~\cite{spatialvlm,3dllm,leo}. See Appendix~\ref{app:spatial} for details.

\section{Experiments}
\label{sec:experiments}

We evaluate TrianguLang on multi-view semantic segmentation across five benchmarks, comparing against click-based and optimization-based methods.

\subsection{Experimental Setup}

\paragraph{Datasets.}
We evaluate on five multi-view benchmarks:
\begin{itemize}[nosep,leftmargin=*]
    \item \textbf{ScanNet++}~\cite{scannet++}: Indoor scenes with dense semantic annotations. We train on the \texttt{nvs\_sem\_train} split (230 scenes) and evaluate on a separate held-out \texttt{nvs\_sem\_val} split (50 scenes).
    \item \textbf{uCO3D}~\cite{UCo3D}: Object-centric 360$^\circ$ captures across 1,000 categories.
    \item \textbf{LERF-OVS}~\cite{lerf}: 4 tabletop scenes with per-frame language-grounded annotations for open-vocabulary segmentation and localization.
    \item \textbf{NVOS}~\cite{NVOS}: 6 multi-view object selection scenes.
    \item \textbf{SPIn-NeRF}~\cite{spinnerf}: 10 real-world multi-view scenes.
\end{itemize}

% Figure moved to appendix for space
\begin{figure}[h]
\begin{center}
\includegraphics[width=\linewidth]
{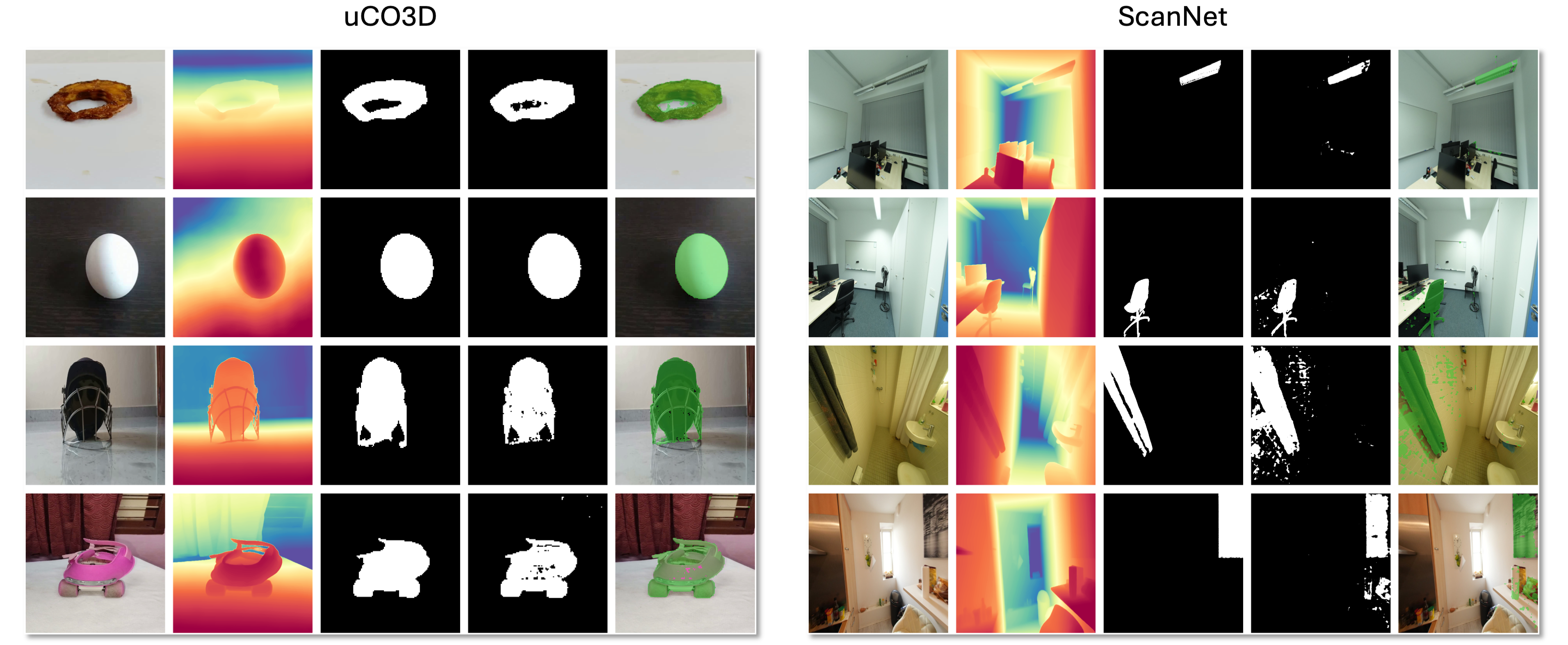}
\end{center}
\caption{Performance on uCO3D and ScanNet++ datasets. Left-to-right: RGB, depth map, ground truth, TrianguLang masks}
\label{fig:grid_uco3d}
\end{figure}

\paragraph{Protocol.}
Following MV-SAM~\cite{mvsam}, we evaluate on 100 frames per scene with 5 randomly sampled objects (excluding structural classes: wall, floor, ceiling). We report mIoU (mean Intersection-over-Union) and mAcc (mean per-class accuracy), averaged across 5 random seeds for object sampling.
\paragraph{Baselines.}
We compare against click-based methods: MV-SAM~\cite{mvsam} (12 grid-sampled point prompts per object, following their published protocol), SAM2-Video~\cite{sam2}, and SAM3~\cite{sam3}; and per-scene optimization methods on NVOS/SPIn-NeRF: SA3D~\cite{SA3D}, SAGA~\cite{SAGA}, OmniSeg3D~\cite{omniseg3d} (using ground-truth 3D reconstructions and calibrated poses). Note that per-scene optimization methods (LERF~\cite{lerf}, LangSplat~\cite{langsplat}) achieve strong accuracy but require calibrated camera poses and 10 to 45 minutes of per-scene training, placing them in a different efficiency regime.

\paragraph{Training.}
Full-scale models train on 230 ScanNet++ scenes (8 views, 100 epochs) or 8,000 uCO3D scenes on 8$\times$ A100 80GB GPUs with DDP. We use AdamW (lr $= 10^{-4}$, cosine schedule, 2-epoch warmup), batch size 8, AMP with FP16, and confidence-based mask selection via align loss~\cite{align2024} at inference.

\subsection{Main Results}

Table~\ref{tab:main} presents in-domain and cross-evaluation results on ScanNet++ and uCO3D. Despite using a single text query instead of 12 click prompts, TrianguLang outperforms all feed-forward baselines in both in-domain and cross-domain settings.

\begin{table}[t]
\caption{In-domain, cross-dataset, and large-scale evaluation results. MV-SAM numbers from~\cite{mvsam}. TrianguLang uses text-only prompts; MV-SAM uses 12 click prompts per object. SA-1B denotes MV-SAM trained on single-view SA-1B data (millions of images).}
\label{tab:main}
\begin{center}
\resizebox{\columnwidth}{!}{
\begin{tabular}{lllcc}
\toprule
Model & Setting & Train $\rightarrow$ Eval & mIoU ($\uparrow$) & mAcc ($\uparrow$) \\
\midrule
\multirow{6}{*}{MV-SAM}
    & In-domain    & ScanNet++ $\rightarrow$ ScanNet++ & 0.510 & 0.694 \\
    & In-domain    & uCo3D $\rightarrow$ uCo3D       & 0.910 & 0.965 \\
    & Cross-domain & uCo3D $\rightarrow$ ScanNet++ & 0.194 & 0.251 \\
    & Cross-domain & ScanNet++ $\rightarrow$ uCo3D   & 0.322 & 0.517 \\
    & Large-scale  & SA-1B $\rightarrow$ ScanNet++    & 0.489 & 0.635 \\
    & Large-scale  & SA-1B $\rightarrow$ uCo3D        & 0.877 & 0.950 \\
\midrule
\multirow{4}{*}{\textbf{TrianguLang (Ours)}}
    & In-domain    & ScanNet++ $\rightarrow$ ScanNet++ & \textbf{0.624} & \textbf{0.774} \\ % v10_full_scale_ep97_seq_proc2; mean_class_recall used as mAcc
    & In-domain    & uCo3D $\rightarrow$ uCo3D       & \textbf{0.946} & \textbf{0.983} \\
    & Cross-domain & uCo3D $\rightarrow$ ScanNet++ & \textbf{0.279} & \textbf{0.685} \\
    & Cross-domain & ScanNet++ $\rightarrow$ uCo3D   & \textbf{0.757} & \textbf{0.796} \\ % v10_full_scale_ep97_uco3d; mean_class_recall
\bottomrule
\end{tabular}
}
\end{center}
\end{table}

\paragraph{In-domain performance.}
On ScanNet++, TrianguLang achieves 62.4\% mIoU with text-only prompts, surpassing MV-SAM's 51.0\% with 12 clicks by 11.4 points (Table~\ref{tab:main}). On uCO3D, the gap is 3.6 points (94.6\% vs.\ 91.0\%). Notably, TrianguLang trained on only 230 ScanNet++ scenes exceeds even MV-SAM trained on the large-scale SA-1B dataset (millions of images) on ScanNet++ evaluation (62.4\% vs.\ 48.9\%), demonstrating that geometric cross-view reasoning compensates for orders of magnitude less training data. On uCO3D, MV-SAM's SA-1B variant (87.7\%) benefits from its massive single-view training corpus, but TrianguLang trained in-domain on uCO3D still surpasses it (94.6\% vs.\ 87.7\%).

\textbf{Disentangling backbone from GASA.} SAM3 alone (no cross-view reasoning) achieves only 48.6\% mIoU, \emph{lower} than MV-SAM's 48.9\% (Table~\ref{tab:benchmark_results} in Appendix). TrianguLang's improvement over SAM3-alone confirms that GASA and world-space PE (not the backbone) drive the gains, as quantified in Table~\ref{tab:ablation}.

\paragraph{Oracle vs.\ predicted IoU.}
For transparency, we report oracle mIoU (selecting the GT-best mask among candidates) alongside predicted mIoU. On ScanNet++ (49 val scenes, 100 frames), SAM3 exhibits a \textbf{30-point} oracle-predicted gap (oracle: 79.9\% vs.\ predicted: 49.8\%), because it generates hundreds of mask candidates per scene and struggles to reliably identify the best one. TrianguLang's gap is only \textbf{1 point} (oracle: 63.4\% vs.\ predicted: 62.4\%), demonstrating that our focused 10-query design virtually eliminates mask selection as a bottleneck. Although SAM3's oracle mIoU (79.9\%) substantially exceeds TrianguLang's (63.4\%), its inability to rank candidates correctly results in actual performance 12.6 points \emph{below} TrianguLang.

\paragraph{Cross-domain generalization.}
Cross-dataset transfer reveals the most significant advantage. TrianguLang trained on ScanNet++ achieves 75.7\% mIoU on uCO3D, more than doubling MV-SAM's 32.2\% (+43.5 points). In the reverse direction (uCO3D$\to$ScanNet++), TrianguLang achieves 27.9\% vs.\ 19.4\% (+8.5 points). The asymmetric transfer suggests that diverse indoor viewpoints and occlusion patterns provide more generalizable geometric priors than single-object captures.

\subsection{Multi-View Benchmarks}

% On NVOS and SPIn-NeRF, TrianguLang achieves 93.5\% and 91.4\% mIoU respectively, surpassing MV-SAM (92.1\% / 92.5\%) among feed-forward methods while using text instead of clicks. TrianguLang approaches per-scene optimization methods (SAGA: 92.6\% / 93.7\%) without any per-scene computation; full per-scene results appear in Appendix~\ref{app:benchmarks}.

On NVOS, TrianguLang achieves 93.5\% mIoU, surpassing MV-SAM (92.1\%) among feed-forward methods while using text instead of clicks. TrianguLang approaches per-scene optimization methods (SAGA: 92.6\% / 93.7\%) without any per-scene computation; full per-scene results appear in Appendix~\ref{app:benchmarks}.

Tables~\ref{tab:lerf_comparison} and~\ref{tab:lerf_detailed} compare TrianguLang against per-scene optimization methods that embed language features into 3D scene representations on the LERF-OVS benchmark. These methods require calibrated camera poses, 3D reconstruction, and minutes of per-scene training. TrianguLang operates feed-forward without any per-scene computation, representing a fundamentally different efficiency/accuracy tradeoff.

\begin{table}[t]
\centering
\caption{Comparison with per-scene language grounding methods on the LERF-OVS benchmark. Per-scene methods require calibrated poses and scene reconstruction. TrianguLang operates feed-forward using ScanNet++-trained weights without any training on LERF scenes.}
\label{tab:lerf_comparison}
\small
\begin{tabular}{@{}lcccc@{}}
\toprule
\textbf{Method} & \textbf{mIoU} & \textbf{Loc. Acc.} & \textbf{Per-scene} & \textbf{Time} \\
\midrule
LERF~\cite{lerf} & 37.4 & 73.6 & \ding{51} & $\sim$45 min \\
LangSplat~\cite{langsplat} & 51.4 & \underline{84.3} & \ding{51} & $\sim$10 min \\
LangSplat-V2~\cite{langsplatv2} & \textbf{59.9} & 84.1 & \ding{51} & $\sim$10 min \\
\midrule
\textbf{TrianguLang}$^\dagger$ & \underline{59.2} & \textbf{89.1} & \ding{55} & $\sim$58ms \\
\bottomrule
\end{tabular}
\\[1mm]
{\footnotesize $^\dagger$Zero-shot transfer from ScanNet++ (no LERF training). Best in \textbf{bold}, second-best \underline{underlined}.}
\vspace{-2mm}
\end{table}

Without any training on LERF data, TrianguLang achieves 59.2\% mIoU and 89.1\% localization accuracy on LERF-OVS, matching LangSplat-V2 on segmentation quality (59.9\% mIoU) while surpassing all methods on localization accuracy ($+$5.0 points over LangSplat-V2's 84.1\%). TrianguLang achieves this three orders of magnitude faster ($\sim$58ms vs.\ 10 to 45 min) and without any per-scene optimization or calibrated camera poses.

\begin{figure}[t]
\centering
\includegraphics[width=\columnwidth]{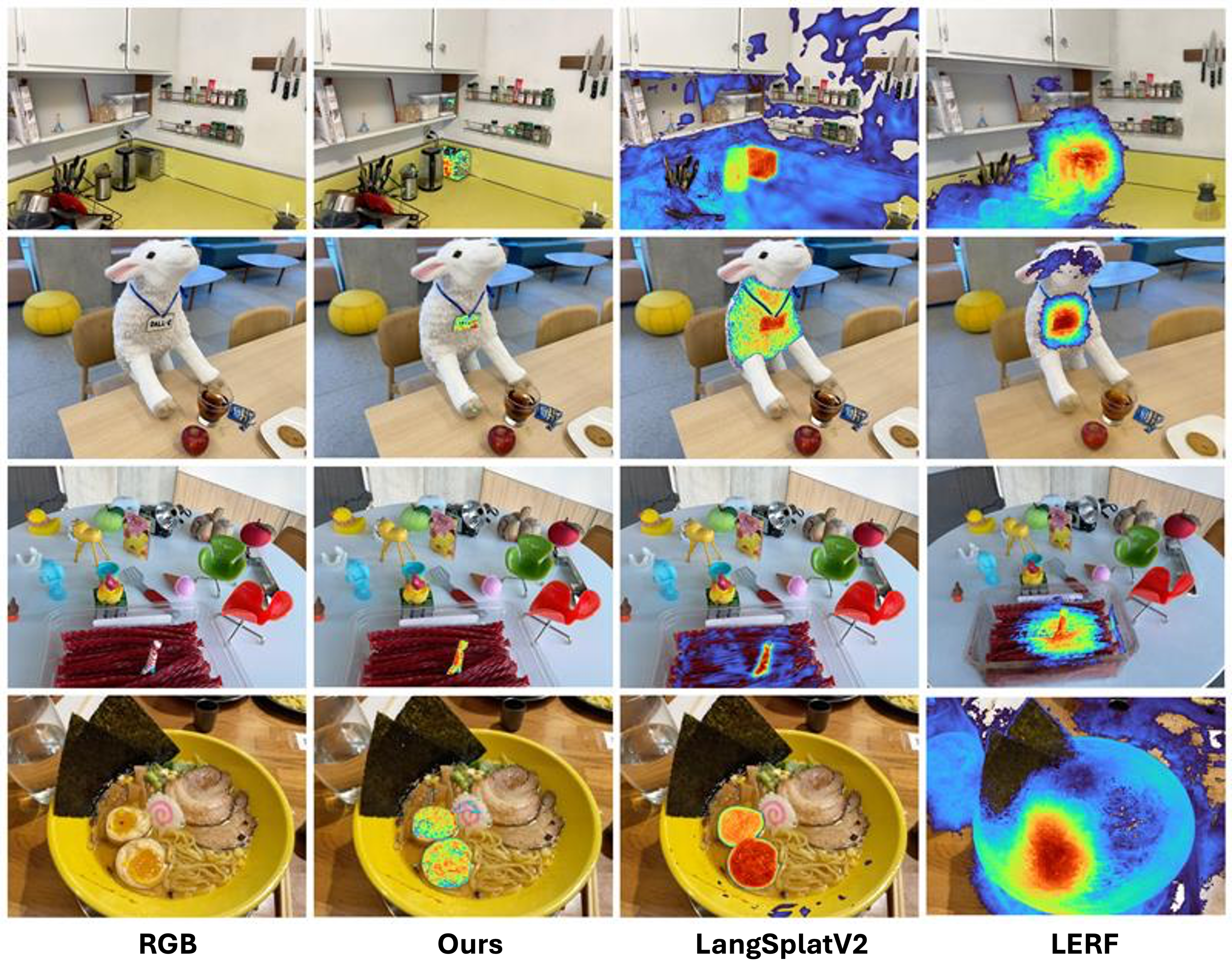}
\caption{Qualitative comparison on LERF-OVS scenes using uniform clip thresholds. Row 1: ``toaster'' query: LERF and LangSplatV2 produce diffuse activations across the scene while TrianguLang tightly focuses its relevancy map on the target object. Row 3: ``stripes'' query: TrianguLang achieves precise localization despite not training on this dataset, and runs 3 orders of magnitude faster ($\sim$58ms vs.\ 10 to 45 min).}
\label{fig:lerf_comparison}
\end{figure}

\begin{table}[h]
\centering
\caption{Per-scene comparison on LERF-OVS benchmark. TrianguLang uses ScanNet++-trained weights without any LERF training (zero-shot transfer). Per-scene methods require 10 to 45 minutes of optimization per scene.}
\label{tab:lerf_detailed}
\resizebox{\linewidth}{!}{
\begin{tabular}{l|ccccc|ccccc}
\toprule
\multirow{2}{*}{\textbf{Method}} & \multicolumn{5}{c|}{\textbf{Localization Accuracy (\%)}} & \multicolumn{5}{c}{\textbf{Semantic Segmentation (mIoU \%)}} \\
 & Ramen & Teatime & Kitchen & Figurines & Overall & Ramen & Teatime & Kitchen & Figurines & Overall \\ \midrule
Gaussian Grouping~\cite{gaussiangrouping} & 32.4 & 69.5 & 50.0 & 44.6 & 49.1 & 26.4 & 54.0 & 31.3 & 34.6 & 36.6 \\
LEGaussians~\cite{legaussians} & 69.0 & 79.7 & 63.6 & 57.1 & 67.4 & 20.2 & 32.3 & 22.3 & 23.4 & 24.6 \\
LERF~\cite{lerf} & 62.0 & 84.8 & 72.7 & 75.0 & 73.6 & 28.2 & 45.0 & 37.9 & 38.6 & 37.4 \\
GOI~\cite{goi} & 56.3 & 67.8 & 68.2 & 44.6 & 59.2 & 33.7 & 55.8 & 54.5 & 23.9 & 42.0 \\
GAGS~\cite{gags} & 69.0 & 88.1 & \underline{90.9} & 78.6 & 81.7 & 46.8 & 60.3 & 55.8 & 53.6 & 54.1 \\
LangSplat~\cite{langsplat} & 73.2 & 88.1 & \textbf{95.5} & 80.4 & \underline{84.3} & \underline{51.2} & \underline{65.1} & 44.5 & 44.7 & 51.4 \\
\midrule
LangSplat-V2~\cite{langsplatv2} & \underline{74.7} & \textbf{93.2} & 86.4 & \underline{82.1} & 84.1 & \textbf{51.8} & \textbf{72.2} & \underline{59.1} & \underline{56.4} & \textbf{59.9} \\
\midrule
\textbf{TrianguLang}$^\dagger$ & \textbf{85.9} & \underline{92.8} & \underline{88.6} & \textbf{98.8} & \textbf{89.1} & 51.1 & 58.9 & \textbf{62.4} & \textbf{62.1} & \underline{59.2} \\
\bottomrule
\end{tabular}}
\\[1mm]
{\footnotesize $^\dagger$Zero-shot transfer from ScanNet++ (no LERF training). Best in \textbf{bold}, second-best \underline{underlined}.}
\end{table}

\subsection{Ablation Studies}

We ablate key components using 230 training scenes with 8 views per scene and sequential sampling, evaluating with matched sampling on 50 held-out scenes (100 frames). Table~\ref{tab:ablation} reports mIoU with changes measured against the full baseline.

\begin{table}[t]
\centering
\caption{Component ablation on ScanNet++ (230 scenes, 8 views, 504 resolution). $\Delta$ measured against ablation baseline (54.7\% mIoU). The full model in Table~\ref{tab:main} achieves 62.4\% at 1008 resolution with 230 scenes.}
\label{tab:ablation}
\small
\begin{tabular}{@{}lcc@{}}
\toprule
\textbf{Configuration} & \textbf{mIoU} & \textbf{$\Delta$} \\
\midrule
Baseline (GASA + World PE) & \textbf{54.7} & $-$ \\
\quad $-$ GASA kernel (standard attn) & 49.4 & $-$5.3 \\
\quad $-$ World-space PE & 49.3 & $-$5.4 \\
\quad $-$ Both (no GASA, no PE) & 46.3 & $-$8.4 \\
\quad RBF kernel (vs.\ learned MLP) & 44.0 & $-$10.7 \\
\bottomrule
\end{tabular}
\vspace{-2mm}
\end{table}

\paragraph{Component contributions.}
Removing the GASA kernel ($-$5.3\%) eliminates the geometric distance bias that suppresses spurious cross-token attention. Without world-space PE ($-$5.4\%), the decoder cannot reason about 3D positions. Removing both components ($-$8.4\%) confirms they are individually critical yet partially complementary: the combined drop is less than the sum of individual drops ($10.7\%$), suggesting shared geometric information.

\paragraph{Visual geometry backbone.}
GASA derives its 3D pointmaps from a frozen visual geometry model (VGM). Table~\ref{tab:vgm_ablation} evaluates three VGMs while keeping the GASA decoder fixed: DA3-NESTED~\cite{da3}, MapAnything~\cite{mapanything}, and $\pi^3$-VO~\cite{pi3}. All three produce multi-view consistent depth and camera poses in a single feed-forward pass without requiring calibration.

\begin{table}[t]
\centering
\caption{Sensitivity to visual geometry backbone. All models use the same GASA decoder architecture trained on ScanNet++ and evaluated zero-shot on uCO3D and LERF-OVS. mIoU (\%) reported.}
\label{tab:vgm_ablation}
\small
\begin{tabular}{@{}lccc@{}}
\toprule
\textbf{VGM} & \textbf{ScanNet++} & \textbf{uCO3D} & \textbf{LERF-OVS} \\
\midrule
DA3-NESTED~\cite{da3}    & 62.4 & \textbf{72.1} & \textbf{59.2} \\
MapAnything~\cite{mapanything}  & \textbf{63.0} & \textbf{72.1} & 58.1 \\
$\pi^3$-VO~\cite{pi3}        & 62.7 & 66.8 & 51.8 \\
\bottomrule
\end{tabular}
\vspace{-2mm}
\end{table}

TrianguLang is robust to VGM choice on in-domain data: all three backbones achieve 62--63\% mIoU on ScanNet++. DA3-NESTED and MapAnything perform comparably across all benchmarks, with MapAnything slightly ahead on ScanNet++ ($+$0.6\%) and DA3-NESTED slightly ahead on LERF-OVS ($+$1.1\%). $\pi^3$-VO lags on the cross-domain benchmarks ($-$5.3\% on uCO3D, $-$7.4\% on LERF-OVS), likely due to less accurate metric depth on out-of-distribution scenes. These results confirm that GASA's geometric reasoning transfers across VGM backends, and suggest that improving the underlying depth model directly benefits downstream segmentation quality.

\subsection{Efficiency}
TrianguLang processes each frame at 1008$\times$1008 resolution in $\sim$58ms on a single A100 GPU, achieving $\sim$17 FPS. With cached depth, preprocessing adds $\sim$32ms per frame ($\sim$90ms total, $\sim$11 FPS). The GASA decoder adds only 13.7M trainable parameters (0.54\% of the 2.54B total), with SAM3 (848M) and DA3-NESTED (1.4B) remaining frozen. For comparison, per-scene optimization methods (SA3D, SAGA, OmniSeg3D) require 30 to 60 minutes of training per new scene; MV-SAM requires 5.1s preprocessing plus 1.1s inference with click prompts. TrianguLang eliminates both per-scene optimization and manual click annotation.

\subsection{Limitations}

\textbf{Depth as bottleneck:} DA3's depth estimates fail on reflective surfaces (mirrors, glass), causing GASA to incorrectly suppress valid correspondences. \textbf{Limited views:} Objects visible in $<$3 views degrade to per-view quality. \textbf{Inherited limitations:} The model inherits SAM3's text-image alignment limitations for rare categories and DA3's metric accuracy ($\sim$5cm error) bounds 3D localization precision. \textbf{Pose estimation on turntable sequences:} DA3-NESTED's pose estimation becomes ill-conditioned on turntable-style captures (e.g., uCO3D) where near-circular camera paths with small baselines cause scale/translation estimation to degenerate; we therefore report only mIoU for uCO3D without Procrustes-aligned 3D localization metrics (ScanNet++ Procrustes alignment remains valid at $\sim$3.9cm median error). \textbf{Training data scale:} TrianguLang was trained on only 230 ScanNet++ scenes, yet outperforms models trained on SA-1B (millions of images); scaling to larger and more diverse training sets would likely amplify the advantage (see Section~\ref{sec:conclusion}).

\section{Conclusion}
\label{sec:conclusion}
We present TrianguLang, a feed-forward framework for language-guided multi-view segmentation and 3D localization that requires no camera calibration or per-scene optimization at inference. Geometry-Aware Semantic Attention (GASA) injects depth-derived geometric bias into cross-view attention, suppressing semantically plausible but geometrically inconsistent correspondences.

Our experiments demonstrate three key findings. First, text-only prompts can surpass click-based methods when combined with geometric cross-view reasoning. Second, the geometric priors learned through GASA transfer strongly across domains: cross-dataset performance more than doubles the best click-based baseline on out-of-domain benchmarks (75.7\% vs.\ 32.2\% on ScanNet++$\to$uCO3D). Third, TrianguLang matches per-scene optimization methods on NVOS and SPIn-NeRF without requiring camera poses, 3D reconstruction, or minutes of per-scene training.

% These results suggest that composing frozen foundation models with lightweight geometric attention is a viable alternative to per-scene optimization. 

Future work includes several directions. \textbf{Scaling training data:} TrianguLang trained on only 230 indoor scenes already surpasses MV-SAM trained on the SA-1B dataset (millions of images) on ScanNet++ evaluation (62.4\% vs.\ 48.9\%), suggesting substantial headroom for scaling with larger and more diverse training corpora. The trend of geometric priors providing disproportionate gains at smaller data scales would likely be amplified with additional data. \textbf{Outdoor and open-world deployment:} Extending to outdoor environments \cite{dl3dv} would enable applications such as language-guided geological survey (e.g., segmenting rock formations by type) and text-only spatial reasoning in unstructured scenes without any visual prompts. Additional future directions include temporal consistency for streaming video, combining learned Spatial IDs~\cite{spatialids2026} with metric depth for improved disambiguation, and deploying TrianguLang on a UR10e robot arm for language-guided manipulation tasks.

% Checklist for ECCV formatting correctness
%\begin{itemize}
%\item I have removed all \verb| \vspace| and \verb|\hspace|  commands from my paper.
%\item I have not used \verb|\cite| command in the abstract.
%\item I have entered a correct \verb|\titlerunning{}| command and selected a meaningful short name for the paper.
%\item I have used the same name spelling in all my papers accepted to ECCV and ECCV Workshops.
%\item I have added acknowledgments without a section number, e.g. using the \verb|\section*%{}| command.
%\item Excluding references and acknowledgments, my paper is no longer than 14 pages.
%\item I have not decreased the font size of any part of the paper (except tables) to fit into 14 pages, I understand Springer editors will remove such commands.
%\end{itemize}

%\clearpage  % TODO FINAL: This \clearpage needs to be removed from both review and camera-ready versions.

%\section*{Acknowledgements}
%Please insert your acknowledgments here.

\bibliographystyle{splncs04}
\bibliography{references}

% TrianguLang APPENDIX

\appendix
\section{Appendix}
\label{app}
\label{app:architecture}
\subsection{GASA Decoder Layer}

Each of the 6 GASA decoder layers processes queries through four sequential operations:

\begin{enumerate}[nosep,leftmargin=*]
    \item \textbf{Self-Attention}: 3 learnable queries attend to each other via standard multi-head attention (8 heads, dimension 32 per head)
    \item \textbf{Text Cross-Attention}: Queries attend to text embeddings from SAM3's text encoder, applied at \emph{every} layer following SAM3's design
    \item \textbf{GASA Cross-Attention}: Queries attend to encoder memory with geometric bias (Eq.~\ref{eq:gasa})
    \item \textbf{Feed-Forward Network}: Two-layer MLP with GELU activation ($256 \to 2048 \to 256$)
\end{enumerate}

% Each operation uses residual connections and layer normalization.

\subsection{Distance Kernel}

The geometric kernel $\phi$ in GASA is a small MLP:
\begin{equation}
\phi(d) = \mathbf{w}_2^\top \cdot \text{ReLU}(\mathbf{w}_1 d + \mathbf{b}_1) + b_2
\end{equation}
with $\mathbf{w}_1, \mathbf{w}_2 \in \mathbb{R}^{32}$. The input distance $d$ is in meters (from DA3 metric depth), providing a natural scale for indoor scenes. We initialize $b_2 = -1$ to encourage suppression of distant matches. The kernel learns to output strongly negative values for large distances; the output is clamped to $[-10, 0]$ for numerical stability.

\subsection{World-Space Positional Encoding}

Following \cite{nerf}, we encode 3D coordinates with multi-frequency sinusoids:
\begin{equation}
\gamma(\mathbf{p}) = \mathbf{p} \oplus \bigoplus_{i=0}^{L-1} \big[ \sin(2^i\pi\mathbf{p}), \cos(2^i\pi\mathbf{p}) \big]
\end{equation}
with $L=10$ frequency bands (63-dimensional). A two-layer MLP projects to model dimension $D=256$.

% \subsection{DA3 Metric Depth Scaling}

% DA3METRIC-LARGE outputs depth calibrated for focal length 300px. To obtain metric depth:
% \begin{equation}
% \mathbf{D}_{\text{metric}} = \mathbf{D}_{\text{raw}} \cdot \frac{f_{\text{DA3}}}{300}
% \end{equation}
% where $f_{\text{DA3}} = f_{\text{orig}} \cdot (H_{\text{DA3}} / H_{\text{orig}})$ accounts for resolution scaling.

\section{Loss Function Details}
\label{app:loss}

\subsection{Segmentation Loss}

\paragraph{Focal Loss.}
Addresses class imbalance (objects typically occupy $<5\%$ of pixels):
\begin{equation}
\mathcal{L}_{\text{focal}} = -\alpha_t (1 - p_t)^\gamma \log(p_t)
\end{equation}
with $\alpha=0.75$ (foreground weight) and $\gamma=2.0$ (focusing parameter).

\paragraph{Dice Loss.}
Optimizes mask overlap directly:
\begin{equation}
\mathcal{L}_{\text{dice}} = 1 - \frac{2 \sum_u \hat{M}(u) \cdot M(u) + \epsilon}{\sum_u \hat{M}(u) + \sum_u M(u) + \epsilon}
\end{equation}

\subsection{Align Loss}

Following AlignDETR and SAM3, the IoU-aware focal loss modulates confidence by mask quality:
\begin{equation}
\mathcal{L}_{\text{align}} = -t_c \cdot (1-p)^\gamma \cdot \log(p) - (1-t_c) \cdot p^\gamma \cdot \log(1-p)
\end{equation}
with soft target $t_c = e^{-r/\tau} \cdot (p^\alpha \cdot u^{1-\alpha})$, where $u$ is actual IoU, $r$ is rank, $\alpha=0.5$, $\tau=2.0$.

\subsection{Default Loss Weights}

\begin{center}
\small
\begin{tabular}{@{}cccccc@{}}
$\lambda_{\text{focal}}$ & $\lambda_{\text{dice}}$ & $\lambda_{\text{align}}$ & $\lambda_{\text{contrastive}}$ & $\lambda_{\text{centroid}}$ & $\lambda_{\text{presence}}$ \\
\midrule
2.0 & 0.5 & 1.0 & 0.3 & 0.5 & 0.5 \\
\end{tabular}
\end{center}

\section{Spatial Language Understanding}
\label{app:spatial}

TrianguLang supports spatial language queries that disambiguate among multiple instances of the same object class (e.g., ``nearest chair'', ``leftmost monitor''). This is achieved through three complementary mechanisms: (1) learnable spatial token embeddings that condition queries during decoding, (2) GT-aware spatial augmentation during training that ensures correct qualifier assignment, and (3) object-aware spatial selection at inference that uses depth to filter predictions post-hoc.

\subsection{Supported Qualifiers}

Spatial qualifiers are parsed from text prompts via prefix matching (sorted by length to handle multi-word qualifiers like ``second nearest'' before ``nearest''). Each qualifier maps to a learnable embedding index. The full vocabulary comprises 48 recognized phrases across 7 base categories, with ordinal and size-based extensions:

\begin{table}[h]
\centering
\caption{Spatial qualifier vocabulary (48 recognized phrases across 7 base categories). Index 0 is reserved for queries without spatial qualifiers. Ordinal and size-based extensions share learnable embedding indices with their related base qualifiers.}
\small
\begin{tabular}{@{}llll@{}}
\toprule
\textbf{Category} & \textbf{Qualifier} & \textbf{Synonyms} & \textbf{Selection Criterion} \\
\midrule
\multicolumn{4}{l}{\textit{Base qualifiers (7 learnable embeddings)}} \\
Depth & nearest & closest, close & $\arg\min_i d_i$ \\
      & farthest & far, distant & $\arg\max_i d_i$ \\
Horizontal & leftmost & left & $\arg\min_i x_i$ \\
           & rightmost & right & $\arg\max_i x_i$ \\
Vertical & topmost & top, upper & $\arg\min_i y_i$ \\
         & bottommost & bottom, lower & $\arg\max_i y_i$ \\
Center & center & central, middle & $\arg\min_i \|c_i - \bar{c}\|$ \\
\midrule
\multicolumn{4}{l}{\textit{Ordinal extensions (share embeddings with base)}} \\
Depth & second nearest & second closest & 2nd-ranked by $d_i$ \\
      & second farthest & & 2nd-ranked by $d_i$ (desc.) \\
Horizontal & second leftmost & second from left & 2nd-ranked by $x_i$ \\
           & second rightmost & second from right & 2nd-ranked by $x_i$ (desc.) \\
Vertical & second topmost & second from top & 2nd-ranked by $y_i$ \\
         & second bottommost & second from bottom & 2nd-ranked by $y_i$ (desc.) \\
Depth & mid-depth & middle depth & median by $d_i$ \\
\midrule
\multicolumn{4}{l}{\textit{Size-based extensions (use mask pixel count)}} \\
Size & largest & biggest, big & $\arg\max_i |M_i|$ \\
     & smallest & small, tiny & $\arg\min_i |M_i|$ \\
\bottomrule
\end{tabular}
\end{table}

\subsection{Spatial Token Embeddings}

Spatial qualifiers are represented as learnable embeddings $\mathbf{E}_{\text{spatial}} \in \mathbb{R}^{8 \times D}$ (initialized with $\mathcal{N}(0, 0.02)$). Given a parsed qualifier index $k \in \{0, \ldots, 7\}$, the spatial embedding is added to all object queries:
\begin{equation}
\mathbf{q}_i \leftarrow \mathbf{q}_i + \mathbf{E}_{\text{spatial}}[k], \quad \forall i \in \{1, \ldots, Q\}
\end{equation}
This broadcast addition biases all queries toward objects matching the spatial criterion, while text scoring and GASA attention still determine the final mask selection.

\subsection{GT-Aware Spatial Augmentation}
\label{app:spatial_gtaware}

A key challenge is training spatial tokens with \emph{correct} qualifier labels. Random augmentation (assigning arbitrary qualifiers) provides no learning signal since the qualifier may not match the target instance. We introduce GT-aware augmentation that verifies spatial qualifiers against ground-truth masks.

\paragraph{Process.}
For each training sample with target object mask $M_{\text{target}}$ and depth map $D$:
\begin{enumerate}[nosep,leftmargin=*]
    \item \textbf{Compute spatial context:} For each visible instance of the target class, compute the 2D centroid $(c_x, c_y)$ (normalized to $[0,1]$) and depth $d$ at the centroid from the DA3 depth map.
    \item \textbf{Determine true qualifiers:} Compare the target instance against all same-class instances. A qualifier is valid only if it is \emph{true}: e.g., ``nearest'' requires $d_{\text{target}} \leq \min(d_{\text{others}}) + \epsilon$.
    \item \textbf{Augment with probability $p$:} With probability $p=0.3$, randomly select one valid qualifier and prepend it to the text prompt (e.g., ``chair'' $\to$ ``nearest chair'').
\end{enumerate}

This ensures the model only trains on correct spatial associations, preventing confusion from false labels.

\paragraph{Multi-instance filtering.}
When \texttt{spatial\_multi\_instance\_only} is enabled, augmentation is skipped for single-instance objects (where spatial qualifiers are trivially satisfied and provide no disambiguation signal).

\subsection{Object-Aware Spatial Selection}
\label{app:spatial_objaware}

At inference, we apply post-hoc spatial filtering using depth. Given $Q$ predicted masks and a spatial qualifier, we compute the 2D centroid and depth at the centroid for each mask, then select the mask satisfying the spatial criterion (e.g., minimum depth for ``nearest''). This is zero-cost: it requires no additional training and works with any pre-trained model, leveraging depth already computed for GASA.

\subsection{Relational Queries}

We parse queries of the form ``\textit{target} \textit{relation} \textit{reference}'' (e.g., ``mug to the right of the keyboard'') using 8 regex patterns that handle common phrasing variations (e.g., ``to the right of,'' ``on the right of,'' and ``right of'' all match \textit{right\_of}). Supported relations:
\begin{itemize}[nosep,leftmargin=*]
    \item Positional: \textit{left\_of}, \textit{right\_of}, \textit{above}, \textit{below} (and synonyms: over, under, beneath)
    \item Depth-based: \textit{in\_front\_of}, \textit{behind}
    \item Proximity: \textit{near} (and synonyms: next to, beside, by), \textit{on\_top\_of} (and synonym: on)
\end{itemize}

During training, the GT-aware augmentor generates relational queries by identifying nearby objects of different classes and checking whether spatial relations hold (e.g., ``chair to the right of the table'' when $c_x^{\text{chair}} > c_x^{\text{table}}$). This extends spatial reasoning to single-instance objects that lack same-class comparisons.

\subsection{Spatial Ablation Results}

Table~\ref{tab:spatial_ablation} ablates spatial components on ScanNet++.

\begin{table}[h]
\centering
\caption{Spatial feature ablation on ScanNet++. Object-aware selection achieves the best segmentation quality.}
\label{tab:spatial_ablation}
\small
\begin{tabular}{@{}lc@{}}
\toprule
\textbf{Configuration} & \textbf{mIoU} \\
\midrule
No spatial features & 40.07 \\
Spatial-as-points & 43.54 \\
Object-aware selection & \textbf{47.01} \\
\bottomrule
\end{tabular}
\end{table}

Object-aware spatial selection achieves the best eval mIoU (47.01\%), a +6.94 point improvement over no spatial features. Encoding spatial qualifiers as point embeddings (spatial-as-points) provides a +3.47 point gain, while the full object-aware pipeline with GT-aware augmentation and depth-based post-hoc selection yields the largest improvement.

\section{Training Details}
\label{app:training}

\subsection{Hyperparameters}

Table~\ref{tab:hyperparams} summarizes the hyperparameters used for training TrianguLang.

\begin{table}[h]
\centering
\caption{Training hyperparameters for TrianguLang.}
\label{tab:hyperparams}
\small
\begin{tabular}{@{}lc@{}}
\toprule
\textbf{Hyperparameter} & \textbf{Value} \\
\midrule
\multicolumn{2}{l}{\textit{Data}} \\
\quad Scenes & 230 \\
\quad Views per sample & 8 \\
\quad Image resolution & 1008 \\
\quad DA3 resolution & 1008 \\
\midrule
\multicolumn{2}{l}{\textit{Optimization}} \\
\quad Optimizer & AdamW \\
\quad Learning rate & $1 \times 10^{-4}$ \\
\quad Weight decay & 0.01 \\
\quad LR scheduler & Cosine annealing \\
\quad Warmup epochs & 2 \\
\quad Batch size (per GPU) & 8 \\
\quad Effective batch size & 64 \\
\quad GPUs & 8$\times$ A100 80GB \\
\quad Epochs & 100 \\
\midrule
\multicolumn{2}{l}{\textit{Architecture}} \\
\quad Model dimension $D$ & 256 \\
\quad Attention heads & 8 \\
\quad Decoder layers & 6 \\
\quad Number of queries & 10 \\
\quad FFN dimension & 2048 \\
\quad GASA $\beta$ init & 1.0 \\
\quad GASA kernel dim & 32 \\
\quad Centroid head & True \\
\bottomrule
\end{tabular}
\end{table}

\subsection{DA3 Depth Model}

We use \textbf{DA3-NESTED-GIANT-LARGE} (1.4B parameters)~\cite{da3} as our visual geometry model (VGM). It processes image chunks jointly, providing metric depth, camera intrinsics, and inter-view camera poses from images alone. This model is used at both training and inference time, placing all views in a shared world frame without requiring ground-truth calibration.

\subsection{Parameter Counts}

\begin{table}[h]
\centering
\caption{Model parameter breakdown.}
\label{tab:params}
\small
\begin{tabular}{@{}lrc@{}}
\toprule
\textbf{Component} & \textbf{Parameters} & \textbf{Trainable} \\
\midrule
SAM3 Backbone & 848M & Frozen \\
DA3 Depth (NESTED-GIANT-LARGE) & 1.4B & Frozen \\
GASA Decoder & 13.4M & \cmark \\
\quad Query embeddings & 26K & \cmark \\
\quad Self-attention layers & 1.2M & \cmark \\
\quad Text cross-attention & 1.2M & \cmark \\
\quad GASA cross-attention & 3.8M & \cmark \\
\quad FFN layers & 1.1M & \cmark \\
\quad Mask prediction head & 5.4M & \cmark \\
\quad Centroid head & 66K & \cmark \\
Query projection + World-space PE & 330K & \cmark \\
\midrule
\textbf{Total} & 2.54B & 13.7M (0.54\%) \\
\bottomrule
\end{tabular}
\end{table}

\subsection{Complete Loss Function}

The full training objective combines segmentation, ranking, and localization terms:
\begin{equation}
\mathcal{L} = \mathcal{L}_{\text{seg}} + \mathcal{L}_{\text{rank}} + \mathcal{L}_{\text{loc}}
\label{eq:total_loss_app}
\end{equation}

where each component is defined as:

\paragraph{Segmentation Loss.}
\begin{equation}
\mathcal{L}_{\text{seg}} = \lambda_f \mathcal{L}_{\text{focal}} + \lambda_d \mathcal{L}_{\text{dice}}
\end{equation}
with $\lambda_f = 2.0$ and $\lambda_d = 0.5$. Focal loss addresses class imbalance with $\alpha=0.75$ (foreground weight) and $\gamma=2.0$ (focusing parameter).

\paragraph{Ranking Loss (Align + Contrastive).}
\begin{equation}
\mathcal{L}_{\text{rank}} = \lambda_a \mathcal{L}_{\text{align}} + \lambda_c \mathcal{L}_{\text{contrastive}}
\end{equation}
with $\lambda_a = 1.0$ and $\lambda_c = 0.3$. The align loss follows SAM3/AlignDETR~\cite{align2024}:
\begin{equation}
\mathcal{L}_{\text{align}} = -t_c \cdot (1-p)^\gamma \cdot \log(p) - (1-t_c) \cdot p^\gamma \cdot \log(1-p)
\end{equation}
where the soft target $t_c = e^{-r/\tau} \cdot (p^\alpha \cdot u^{1-\alpha})$ incorporates rank $r$, prediction confidence $p$, and actual IoU $u$, with $\alpha=0.5$ and $\tau=2.0$.

The contrastive loss encourages correct mask ranking:
\begin{equation}
\mathcal{L}_{\text{contrastive}} = \sum_{i,j: u_i > u_j} \max(0, m - (s_i - s_j))
\end{equation}
where $s_i, s_j$ are predicted confidence scores for masks $i, j$ respectively, $u_i, u_j$ are their actual IoU values with the ground truth, and $m=0.5$ is the margin.

\paragraph{Localization Loss.}
\begin{equation}
\mathcal{L}_{\text{loc}} = \lambda_{\text{3d}} \mathcal{L}_{\text{centroid}} + \lambda_p \mathcal{L}_{\text{presence}}
\end{equation}
with $\lambda_{\text{3d}} = 0.5$ and $\lambda_p = 0.5$. Centroid loss uses smooth L1:
\begin{equation}
\mathcal{L}_{\text{centroid}} = \text{SmoothL1}(\hat{\mathbf{c}}, \mathbf{c}_{\text{GT}})
\end{equation}
Presence loss is binary cross-entropy for object existence prediction:
\begin{equation}
\mathcal{L}_{\text{presence}} = -\left[ y \log(\hat{p}) + (1 - y) \log(1 - \hat{p}) \right]
\end{equation}
where $y \in \{0, 1\}$ indicates whether the queried object is visible in the scene and $\hat{p}$ is the model's predicted existence probability, derived from the highest-confidence mask score.

\paragraph{Sheaf Consistency Loss.}
The sheaf loss enforces cross-view agreement with weight $\lambda_s = 0.1$:
\begin{equation}
\mathcal{L}_{\text{sheaf}} = \frac{1}{|\mathcal{C}|} \sum_{(i,j)} \sum_{(\mathbf{u}_i, \mathbf{u}_j) \in \mathcal{C}_{ij}} \left( \sigma(\hat{M}_i(\mathbf{u}_i)) - \sigma(\hat{M}_j(\mathbf{u}_j)) \right)^2
\end{equation}
where $\mathcal{C}_{ij}$ are corresponding pixels with 3D distance $< 5$cm.

\paragraph{Default Loss Weights.}
\begin{center}
\small
\begin{tabular}{@{}cccccc@{}}
$\lambda_f$ & $\lambda_d$ & $\lambda_a$ & $\lambda_c$ & $\lambda_{\text{3d}}$ & $\lambda_s$ \\
\midrule
2.0 & 0.5 & 1.0 & 0.3 & 0.5 & 0.1 \\
\end{tabular}
\end{center}

\section{Ablations and Negative Results}
\label{app:negative}

We document approaches that did \textit{not} improve performance, to aid future research.

% \subsection{Learning Rate Sensitivity}

% We found $\text{lr} = 1 \times 10^{-4}$ to be optimal. Higher learning rates ($1 \times 10^{-3}$) caused training instability with NaN losses, while lower rates ($1 \times 10^{-5}$) led to slow convergence without improvement.

\subsection{Semantic Union vs Instance GT}

For text-only training, we compared two ground truth strategies:
\begin{itemize}[nosep,leftmargin=*]
    \item \textbf{Semantic Union}: Merge all instances of queried class (e.g., all ``chairs'')
    \item \textbf{Instance GT}: Random single instance per class
\end{itemize}

Counter-intuitively, instance GT generalized better despite the apparent mismatch with text semantics. We hypothesize this prevents the model from learning to ``over-segment'' and produces tighter masks.

% \subsection{Sheaf Loss Without Poses}

% Sheaf loss requires world-consistent pointmaps across views. Using per-frame depth (DA3METRIC-LARGE) without camera poses caused the sheaf loss to learn incorrect correspondences, hurting rather than helping cross-view consistency. Sheaf loss should only be enabled with:
% \begin{itemize}[nosep,leftmargin=*]
%     \item Ground-truth camera poses (from dataset), or
%     \item DA3-NESTED estimated poses (consistent across chunk)
% \end{itemize}

\subsection{Number of Queries: 10 vs.\ 50 vs.\ 100}

We ablated query count on 20-scene evaluation. The default 10 queries achieved 38.5\% mIoU; increasing to 50 queries yielded a slight drop to 37.5\% mIoU; and 100 queries caused complete collapse (0.0\% mIoU). While higher query counts improve oracle metrics (selecting the GT-best mask among candidates), the align loss-based confidence scoring becomes progressively unreliable as the candidate pool grows. This progressive degradation is consistent with SAM3's behavior: SAM3 generates hundreds of candidates and achieves an impressive 79.9\% oracle mIoU on ScanNet++, yet its predicted mIoU is only 49.8\%, a 30-point gap caused by the difficulty of identifying the best mask among many plausible candidates. TrianguLang's focused 10-query design strikes the optimal balance between coverage and selectability, achieving only a 1-point oracle-predicted gap (63.4\% vs.\ 62.4\%).

\subsection{Cross-View Attention}

True cross-view attention (concatenating memories from all views) increased training cost 4$\times$ without significant accuracy gains on our benchmarks. This was analyzed in depth in \cite{mvsam}. The per-view processing with world-space PE achieved similar cross-view consistency more efficiently.

\subsection{Oracle vs.\ Predicted IoU Analysis}
\label{app:oracle}

Table~\ref{tab:oracle} compares oracle and predicted mIoU for TrianguLang and SAM3 on ScanNet++. Oracle mIoU selects the mask with highest GT IoU among all candidates, upper-bounding the model's segmentation quality and isolating mask selection as the performance bottleneck. We report these numbers for transparency: while SAM3's architecture can produce high-quality masks (79.9\% oracle), its practical utility is limited by the inability to identify the correct mask among its many proposals.

\begin{table}[h]
\centering
\caption{Oracle vs.\ predicted mIoU on ScanNet++ (49 val scenes, 100 frames/scene). Oracle selects the GT-best mask among candidates. The gap quantifies the mask selection bottleneck.}
\label{tab:oracle}
\small
\begin{tabular}{@{}lcccc@{}}
\toprule
\textbf{Method} & \textbf{Queries} & \textbf{Predicted} & \textbf{Oracle} & \textbf{Gap} \\
\midrule
SAM3~\cite{sam3} & $\sim$200 & 49.8 & \textbf{79.9} & 30.1 \\
\textbf{TrianguLang} & 10 & \textbf{62.4} & 63.4 & \textbf{1.0} \\
\bottomrule
\end{tabular}
\end{table}

TrianguLang's 10-query design produces a gap of only 1.0 point, indicating that the align loss + contrastive ranking effectively calibrates mask confidence among a small candidate pool. SAM3's 30.1-point gap reveals that its hundreds of mask proposals, while individually high-quality, overwhelm the selection mechanism, a fundamental limitation of generate-then-select architectures at high candidate counts.

\section{Dataset Details}
\label{app:datasets}

\subsection{ScanNet++}

We use 230 scenes with semantic annotations from the \texttt{nvs\_sem\_train} split. Each scene contains $\sim$500 images each at 1752$\times$1168 resolution (resized\_undistorted). We sample 4-8 views per training iteration with diverse viewpoint coverage at 1008$\times$1008 resolution.

\paragraph{Label Normalization.}
ScanNet++ contains label inconsistencies (typos, pluralization). We normalize labels via a fixed dictionary mapping.

\paragraph{Rasterization artifacts.}
The 2D instance masks are obtained by rasterizing the 3D mesh annotation onto each frame using camera poses. This process can produce phantom masks where occluded objects ``bleed through'' walls (e.g., jeans visible through a wall from an adjacent room). We do not filter these artifacts, as they are rare ($<$1\% of masks) and affect baselines equally.

\subsection{uCO3D}

uCO3D provides 360$^\circ$ object-centric captures across 1,000 LVIS categories. Each sequence contains $\sim$150 frames at 1080p resolution with instance segmentation masks. We use this for both in-domain training and out-of-domain generalization testing.

\paragraph{Data cleaning.} We identified and excluded several annotation issues in uCO3D:
\begin{itemize}[nosep,leftmargin=*]
    \item \textbf{Mislabeled sequences:} Some sequences contain objects that do not match their category label (e.g., a sequence labeled \texttt{bow\_weapon} actually shows a helmet; a \texttt{power\_shovel} sequence contains a tea kettle). We manually identified and excluded 12 such sequences.
    \item \textbf{Bad mask categories:} Categories with systematically poor mask quality (e.g., hollow objects with incorrectly filled centers) were excluded entirely (3 categories: \texttt{coil}, \texttt{table-tennis\_table}, \texttt{pickle}).
    \item \textbf{Prompt normalization:} LVIS category names (e.g., \texttt{arctic\_type\_of\_shoe}, \texttt{cab\_taxi}) were mapped to common English terms (``shoe,'' ``taxi'') via a 373-entry normalization dictionary to improve text encoder alignment.
\end{itemize}

\subsection{NVOS and SPIn-NeRF}

NVOS contains 6 scenes with multi-view selection annotations. SPIn-NeRF provides 10 real-world scenes. Both test multi-view segmentation without training.

% \subsection{PartImageNet}

% PartImageNet provides part-level annotations for 158 ImageNet categories with 24,000 images. We render these as multi-view sequences using random camera poses around the object mesh.

\section{Inference Without Ground Truth}
\label{app:inference}

A key design goal is \textbf{no ground-truth camera parameters at any stage}, neither training nor inference.

% \paragraph{What requires GT at training vs inference:}

% \begin{center}
% \small
% \begin{tabular}{@{}lcc@{}}
% \toprule
% \textbf{Component} & \textbf{Training} & \textbf{Inference} \\
% \midrule
% Depth estimation & DA3-NESTED estimated & DA3-NESTED estimated \\
% Camera intrinsics & DA3-NESTED estimated & DA3-NESTED estimated \\
% Camera poses & DA3-NESTED estimated & DA3-NESTED estimated \\
% 3D localization & DA3-NESTED estimated & DA3-NESTED estimated \\
% Segmentation GT & Dataset annotations & Not needed \\
% \bottomrule
% \end{tabular}
% \end{center}
% \textbf{Note:} No ground-truth camera poses or intrinsics are required at any stage. DA3-NESTED-GIANT-LARGE jointly estimates metric depth, intrinsics, and inter-view poses from images alone, placing all views in a shared world frame at both training and inference time.

The model outputs 3D centroids in DA3-NESTED's estimated world frame, which is sufficient for robotics and AR applications without requiring external calibration or ground-truth alignment.

\section{Computational Cost}
\label{app:compute}

\begin{table}[h]
\centering
\caption{Training and inference computational cost.}
\label{tab:compute}
\small
\begin{tabular}{@{}lcccc@{}}
\toprule
\textbf{Config} & \textbf{GPUs} & \textbf{Time/epoch} & \textbf{Memory} & \textbf{Inference} \\
\midrule
Full (230s, 8v) & 8$\times$A100 & $\sim$10 min & 45 GB & $\sim$107ms \\
\bottomrule
\end{tabular}
\end{table}

\paragraph{Inference Speed.}
On a single A100, TrianguLang processes each frame at 1008$\times$1008 resolution in $\sim$58ms ($\sim$17 FPS). With preprocessing (cached DA3-NESTED depth loading and image normalization), end-to-end latency is $\sim$90ms ($\sim$11 FPS). Inference includes SAM3 encoding, GASA decoding, and mask prediction.

\section{Extended Benchmark Results}
\label{app:benchmarks}

Table~\ref{tab:nvos_spinnerf} provides detailed comparisons on NVOS and SPIn-NeRF, including per-scene optimization methods that require calibrated camera poses and 30 to 60 minutes of training per scene.

\begin{table}[h]
\centering
\caption{Comparison on NVOS and SPIn-NeRF benchmarks (mIoU). Per-scene optimization methods require calibrated poses and 30 to 60 minutes of training per scene. TrianguLang and MV-SAM are feed-forward methods requiring no per-scene computation.}
\label{tab:nvos_spinnerf}
\small
\begin{tabular}{@{}llcc@{}}
\toprule
\textbf{Type} & \textbf{Method} & \textbf{NVOS} & \textbf{SPIn-NeRF} \\
\midrule
\multirow{3}{*}{Feed-forward} & SAM2-Video & 88.7 & 86.6 \\
& MV-SAM & 92.1 & 92.5 \\
& \textbf{TrianguLang} & \textbf{93.5} & 91.4 \\
\midrule
\multirow{4}{*}{\shortstack[l]{Per-scene\\optimization}} & SPIn-NeRF & $-$ & 90.7 \\
& SA3D & 91.1 & 92.4 \\
& SAGA & 92.6 & 93.7 \\
& OmniSeg3D & 92.8 & \textbf{94.5} \\
\bottomrule
\end{tabular}
\end{table}

\subsection{Extended Baseline Comparison}

Table~\ref{tab:benchmark_results} provides detailed comparison with recent video segmentation baselines, including SAM2-Long~\cite{ding2024sam2long} and SAM3~\cite{sam3}. The ``Prompt Projection'' baseline unprojects user prompts into 3D and re-projects into each view to run SAM2 independently; this naive approach suffers from occlusion issues in complex indoor scenes.

\begin{table}[h]
\centering
\caption{Comparison of mIoU and mAcc on ScanNet++ and uCo3D. All baselines use their respective SA-1B-trained weights. Prompt Projection$^\dagger$ unprojects prompts into 3D and re-projects per view to run SAM2 independently. SAM3 alone (no cross-view reasoning) underperforms MV-SAM, confirming that TrianguLang's gains come from GASA, not the backbone.}
\label{tab:benchmark_results}
\resizebox{\linewidth}{!}{
\begin{tabular}{lcccc}
\toprule
\multirow{2}{*}{Method} &
\multicolumn{2}{c}{ScanNet++} &
\multicolumn{2}{c}{uCo3D} \\
\cmidrule(lr){2-3}
\cmidrule(lr){4-5}
 & mIoU ($\uparrow$) & mAcc ($\uparrow$) & mIoU ($\uparrow$) & mAcc ($\uparrow$) \\
\midrule
Prompt Projection$^\dagger$ & 29.2 & 59.2 & 78.2 & 83.3 \\
SAM2-Video~\cite{sam2} & 46.1 & 61.4 & 81.9 & 91.3 \\
SAM2-Long~\cite{ding2024sam2long} & 41.5 & 61.4 & 72.9 & 86.4 \\
SAM3~\cite{sam3} & 48.6 & 63.4 & 82.4 & 91.4 \\
MV-SAM & 48.9 & 63.5 & 87.7 & 95.0 \\
\midrule
\textbf{TrianguLang (Ours)} & \textbf{62.4} & \textbf{77.4} & \textbf{94.6} & \textbf{98.3} \\
\bottomrule
\end{tabular}
}
\end{table}

\subsection{Qualitative Results on NVOS and SPIn-NeRF}

We present qualitative results on representative NVOS and SPIn-NeRF scenes in Figures~\ref{fig:nvos_trex} and~\ref{fig:spinnerf_table}. TrianguLang produces clean, accurate segmentation masks using only text prompts, successfully handling challenging cases including thin structures (fern leaves), reflective surfaces (fortress windows), and complex skeletal geometry (trex).

On the LERF-OVS benchmark (Figure~\ref{fig:lerf_comparison}), TrianguLang achieves 59.2\% mIoU and 89.1\% localization accuracy without any scene-specific training, closely matching LangSplat-V2 (59.9\% mIoU) while running three orders of magnitude faster. This demonstrates that strong foundation model features combined with geometric cross-view reasoning can compete with methods that explicitly optimize language-3D associations per scene.
% \begin{itemize}[nosep,leftmargin=*]
%     \item SAM3 encoding: 30ms/view
%     \item DA3 depth: 50ms/view
%     \item GASA decoding: 15ms total
%     \item Mask prediction: 5ms/view
% \end{itemize}

\subsection{Per-Scene Results for NVOS and SPIn-NeRF}
\label{subsec:perscene_results}
Since we excluded the \textit{orchid} scene from NVOS and the \textit{pinecone} scene from SPIn-NeRF,
we provide per-scene results for SAM2, MV-SAM, and prior baselines~\cite{spinnerf, SA3D,  SAGA, omniseg3d} to facilitate future research.
Per-scene mIoU and mAcc are reported in Table~\ref{tab:per_scene_nvos} for NVOS and in Table~\ref{tab:per_scene_spinnerf} for SPIn-NeRF.

% -- Table: NVOS Per-scene Results --
\begin{table}[!h]
\centering
\caption{Per-scene quantitative results on NVOS. The first row of each method corresponds to metric mIoU and the second row to metric mAcc. TrianguLang achieves state-of-the-art average mIoU (93.46\%) among feed-forward methods, approaching per-scene optimization methods while using only text prompts and no per-scene training.}
\label{tab:per_scene_nvos}
\resizebox{\linewidth}{!}{
\begin{tabular}{lcccccccc}
\toprule
Method & fern & flower & fortress & horns\_center & horns\_left & leaves & trex & Avg. \\
\midrule
\multirow{2}{*}{SA3D~\cite{SA3D}}
 & 82.90 & 94.60 & 98.30 & 96.20 & 90.20 & 93.20 & 81.99 & 91.06 \\
 & 94.39 & 98.74 & 99.68 & 99.33 & 99.36 & 99.57 & 97.41 & 98.35 \\
\midrule
\multirow{2}{*}{SAGA~\cite{SAGA}}
 & 83.53 & 96.62 & 98.16 & \textbf{98.06} & 93.59 & 93.51 & 80.81 & 92.57 \\
 & 94.60 & \textbf{99.17} & 99.65 & 99.50 & 99.51 & \textbf{99.59} & 97.26 & 98.55 \\
\midrule
\multirow{2}{*}{OmniSeg3D~\cite{omniseg3d}}
 & 82.70 & 95.30 & \textbf{98.50} & 97.70 & 95.60 & 92.70 & \textbf{87.40} & 92.84 \\
 & 94.30 & 98.90 & \textbf{99.70} & \textbf{99.60} & \textbf{99.70} & 99.50 & \textbf{98.30} & \textbf{98.57} \\
\midrule
\multirow{2}{*}{SAM2~\cite{sam2}}
 & 82.83 & 95.20 & 97.03 & 95.71 & 94.66 & 93.37 & 62.24 & 88.72 \\
 & 93.94 & 97.75 & 98.52 & 97.93 & 95.80 & 96.84 & 81.66 & 94.63 \\
\midrule
\multirow{2}{*}{MV-SAM}
 & 82.90 & 95.50 & 97.50 & 97.60 & 94.50 & 94.30 & 82.50 & 92.11 \\
 & 94.90 & 98.30 & 98.70 & 97.50 & 98.40 & 98.50 & 95.90 & 97.46 \\
\midrule
\multirow{2}{*}{\textbf{TrianguLang (Ours)}}
 & \textbf{84.40} & \textbf{98.12} & 97.40 & 97.60 & \textbf{95.80} & \textbf{94.90} & 86.00 & \textbf{93.46} \\
 & \textbf{96.80} & 98.40 & 97.40 & 98.00 & 98.60 & 95.20 & 93.30 & 96.81 \\
\bottomrule
\end{tabular}
}
\end{table}

% -- Table: SPIn-NeRF Per-scene Results --
\begin{table}[!h]
\centering
\caption{Per-scene quantitative results on SPIn-NeRF. The first row of each method corresponds to metric mIoU and the second row to metric mAcc. TrianguLang achieves competitive results (91.35\% mIoU) using only text prompts without any per-scene optimization, approaching per-scene methods like SA3D (92.36\%) while being orders of magnitude faster.}
\label{tab:per_scene_spinnerf}
\resizebox{\linewidth}{!}{
\begin{tabular}{lcccccccccc}
\toprule
Method & room & orchids & horns & fern & fortress & leaves & fork & truck & lego & Avg. \\
\midrule
\multirow{2}{*}{SPIn-NeRF~\cite{spinnerf}}
 & 95.6 & \textbf{92.7} & 92.8 & 94.3 & 97.7 & 94.9 & 87.9 & 85.2 & 74.9 & 90.67 \\
 & 99.4 & \textbf{98.8} & 98.7 & 99.2 & 99.7 & 99.7 & 99.5 & 95.1 & 99.2 & 98.81 \\
\midrule
\multirow{2}{*}{SA3D~\cite{SA3D}}
 & 88.22 & 83.55 & \textbf{94.49} & 97.05 & \textbf{98.33} & \textbf{97.18} & 89.41 & 90.82 & 92.15 & 92.36 \\
 & 98.33 & 96.86 & \textbf{99.02 }& 99.59 & \textbf{99.75} & \textbf{99.85} & 99.55 & 96.66 & 99.75 & 98.82 \\
\midrule

\multirow{2}{*}{SAGA~\cite{SAGA}}
 & 96.91 & 90.55 & 92.96 & 96.49 & 96.16 & 95.52 & 85.84 & 95.71 & \textbf{93.17} & 93.70 \\
 & 99.59 & 98.29 & 98.71 & 99.51 & 99.41 & 99.75 & 99.42 & 98.53 & \textbf{99.79} & 99.22 \\
\midrule
\multirow{2}{*}{OmniSeg3D~\cite{omniseg3d}}
 & \textbf{97.9} & 92.3 & 91.5 & 97.5 & 97.9 & 96.0 & 90.4 & 96.1 & 90.8 & \textbf{94.49} \\
 & \textbf{99.7} & 98.7 & 98.5 & \textbf{99.7} & 99.7 & 99.8 & \textbf{99.6} & \textbf{98.7} & 99.7 & \textbf{99.34} \\
\midrule
\multirow{2}{*}{SAM2-Video~\cite{sam2}}
 & 91.90 & 86.90 & 84.44 & 97.33 & 69.49 & 96.76 & 83.88 & 83.49 & 85.25 & 86.60 \\
 & 96.00 & 94.32 & 92.39 & 98.88 & 84.76 & 98.53 & 92.67 & 91.86 & 93.01 & 93.60 \\
\midrule
\multirow{2}{*}{MV-SAM~\cite{mvsam}}
 & 91.50 & 83.50 & 89.60 & 97.40 & 98.10 & 95.80 & 90.00 & \textbf{96.40} & 89.90 & 92.47 \\
 & 95.90 & 95.70 & 95.20 & 99.00 & 99.10 & 98.10 & 96.00 & 98.60 & 96.90 & 97.17 \\
\midrule
\multirow{2}{*}{\textbf{TrianguLang (Ours)}}
 & 90.7 & 84.6 & 93.45 & \textbf{98.2} & 96.7 & 97.0 & \textbf{93.0} & 83.2 & 85.3 & 91.35 \\
 & 97.2 & 96.1 & 97.4 & 97.9 & 97.9 & 97.2 & 97.2 & 83.3 & 96.1 & 95.59 \\
\bottomrule
\end{tabular}}
\end{table}

\subsection{LERF-OVS Benchmark Results}

Per-scene LERF-OVS results are reported in Table~\ref{tab:lerf_detailed} (main text). TrianguLang was not trained on any LERF data; these results represent zero-shot transfer from our ScanNet++-trained model. TrianguLang achieves 59.2\% overall mIoU, closely matching LangSplat-V2 (59.9\%) while requiring no per-scene optimization. Our method achieves the highest mIoU on Ramen (63.7\%) and Figurines (56.6\%), and the highest localization accuracy on Figurines (96.6\%), demonstrating strong generalization to cluttered tabletop scenes despite training only on room-scale indoor environments.

% \paragraph{Prompt normalization.}

\subsection{Qualitative Results}

\begin{figure}[h]
\centering
\includegraphics[width=\linewidth]{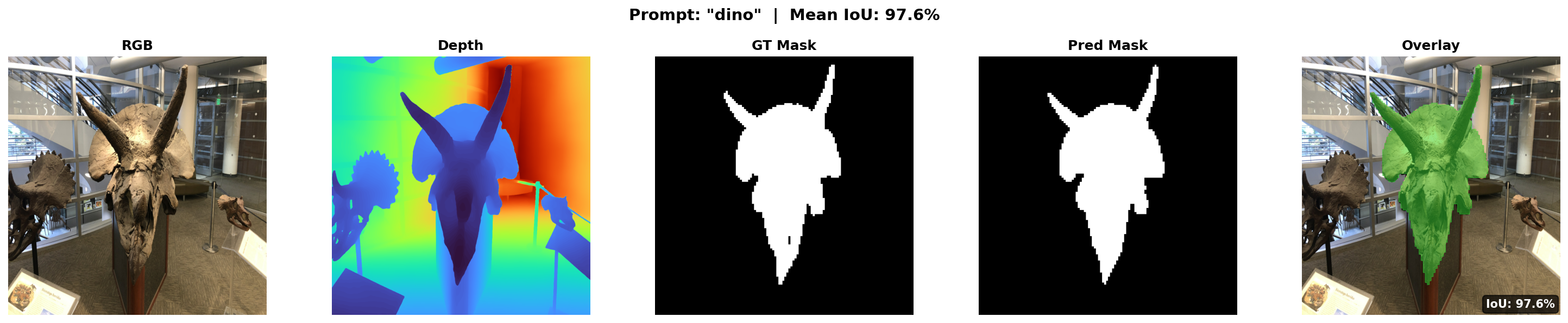}\\[2mm]
\includegraphics[width=\linewidth]{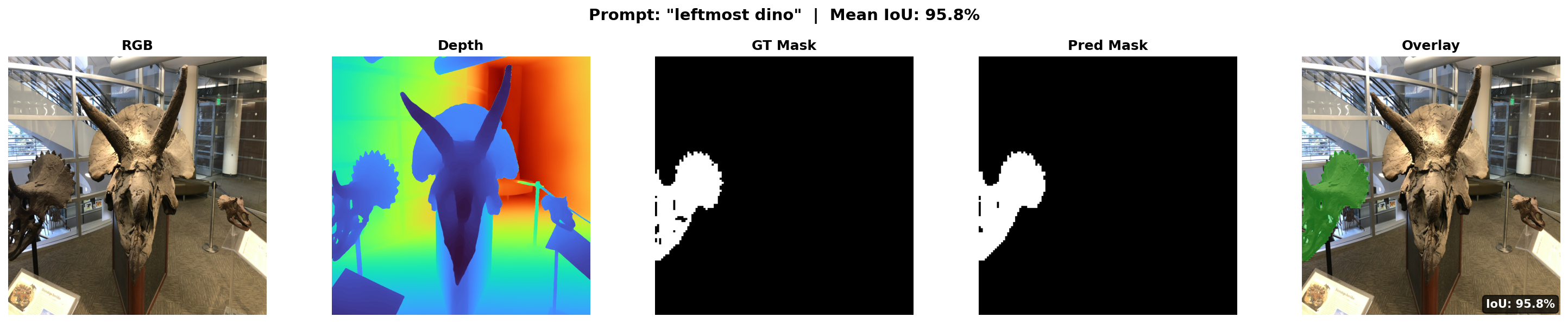}
\caption{Spatial disambiguation on the NVOS T-Rex scene. \textbf{Top:} The query ``dino'' (97.6\% IoU) segments the dominant triceratops skull in the scene. \textbf{Bottom:} The query ``leftmost dino'' (95.8\% IoU) leverages spatial reasoning to disambiguate between the two skulls, correctly selecting only the left specimen. The depth map (second column) provides the geometric context that enables this: TrianguLang computes 3D centroids for each candidate mask and selects the one satisfying the spatial constraint ($\arg\min_i x_i$ for ``leftmost''), resolving ambiguity that would be impossible with object names alone.}
\label{fig:nvos_trex}
\end{figure}

\begin{figure}[h]
\centering
\includegraphics[width=0.9\linewidth]{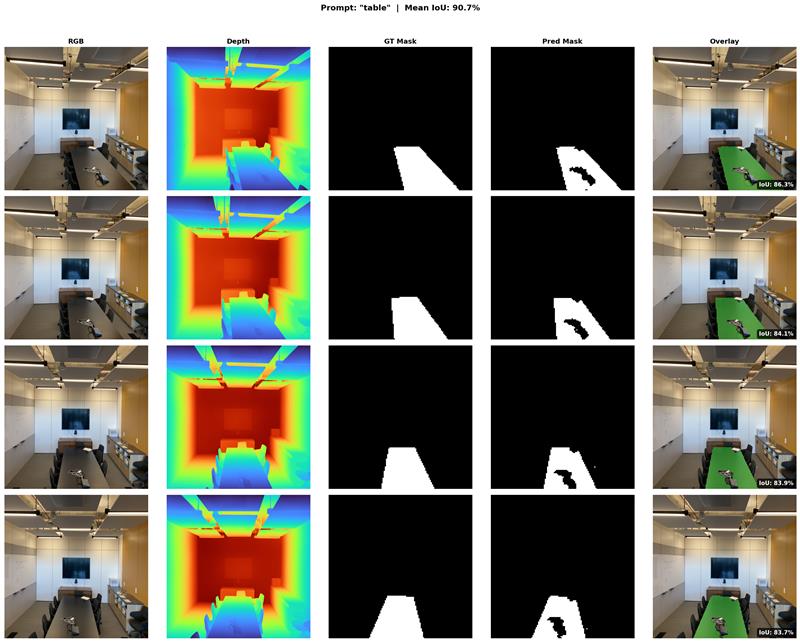}
\caption{Segmentation results on the SPIn-NeRF room scene (90.7\% mean IoU). Each row shows a different viewpoint: RGB input, DA3 depth estimate, ground truth mask, predicted mask, and overlay. When queried for ``table,'' TrianguLang produces a clean segmentation of the table surface \emph{without} including the conference equipment (microphones) present in the ground truth annotation, demonstrating learned semantic boundaries rather than memorized annotation artifacts.}
\label{fig:spinnerf_table}
\end{figure}

Figure~\ref{fig:nvos_trex} demonstrates spatial disambiguation on the NVOS T-Rex scene. When queried with just ``dino,'' the model segments the dominant triceratops skull. However, when the query becomes ``leftmost dino,'' spatial reasoning activates: the model identifies all candidate masks for ``dino,'' computes their 3D centroids via depth unprojection, and selects the one with the minimum horizontal coordinate. This correctly isolates the left skull despite both specimens being semantically similar, an ambiguity that object names alone cannot resolve.

Figure~\ref{fig:spinnerf_table} illustrates a notable capability on the SPIn-NeRF room scene: when queried for ``table,'' TrianguLang produces a clean segmentation of the table surface \emph{without} including the conference equipment (microphones) that are present in the ground truth annotation. This demonstrates that our model learns semantic object boundaries rather than memorizing annotation artifacts.

\subsection{3D Mesh Reconstruction}

\begin{figure}[h]
\centering
\includegraphics[width=0.32\linewidth]{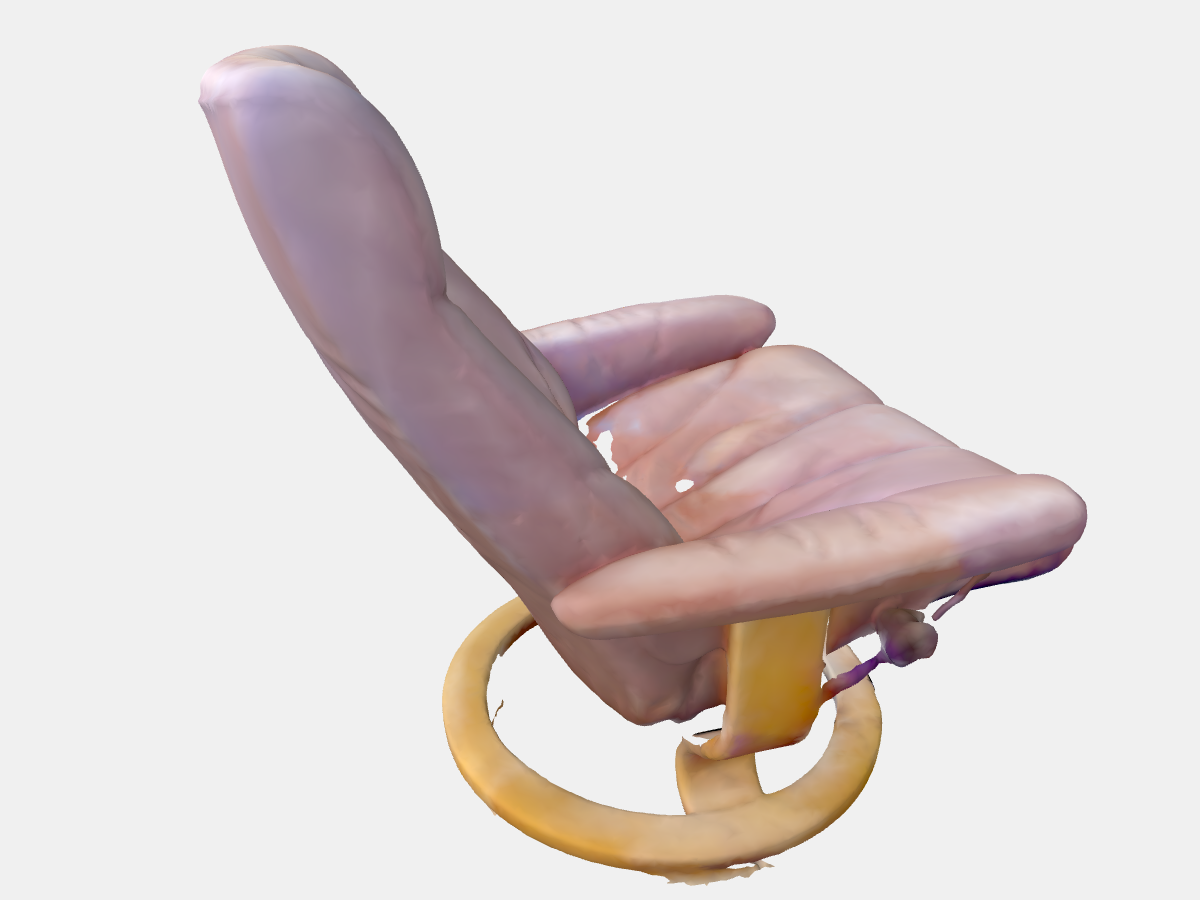}%
\hfill
\includegraphics[width=0.32\linewidth]{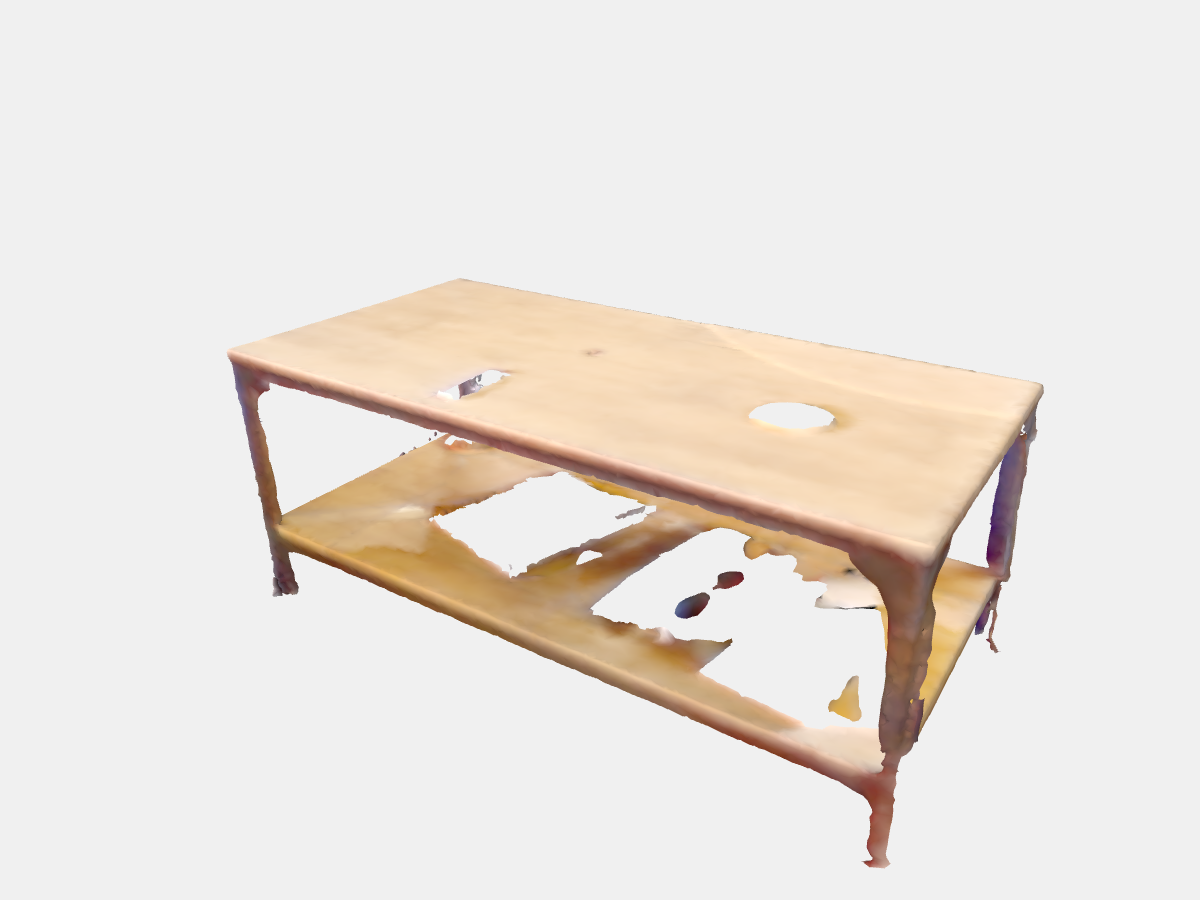}%
\hfill
\includegraphics[width=0.32\linewidth]{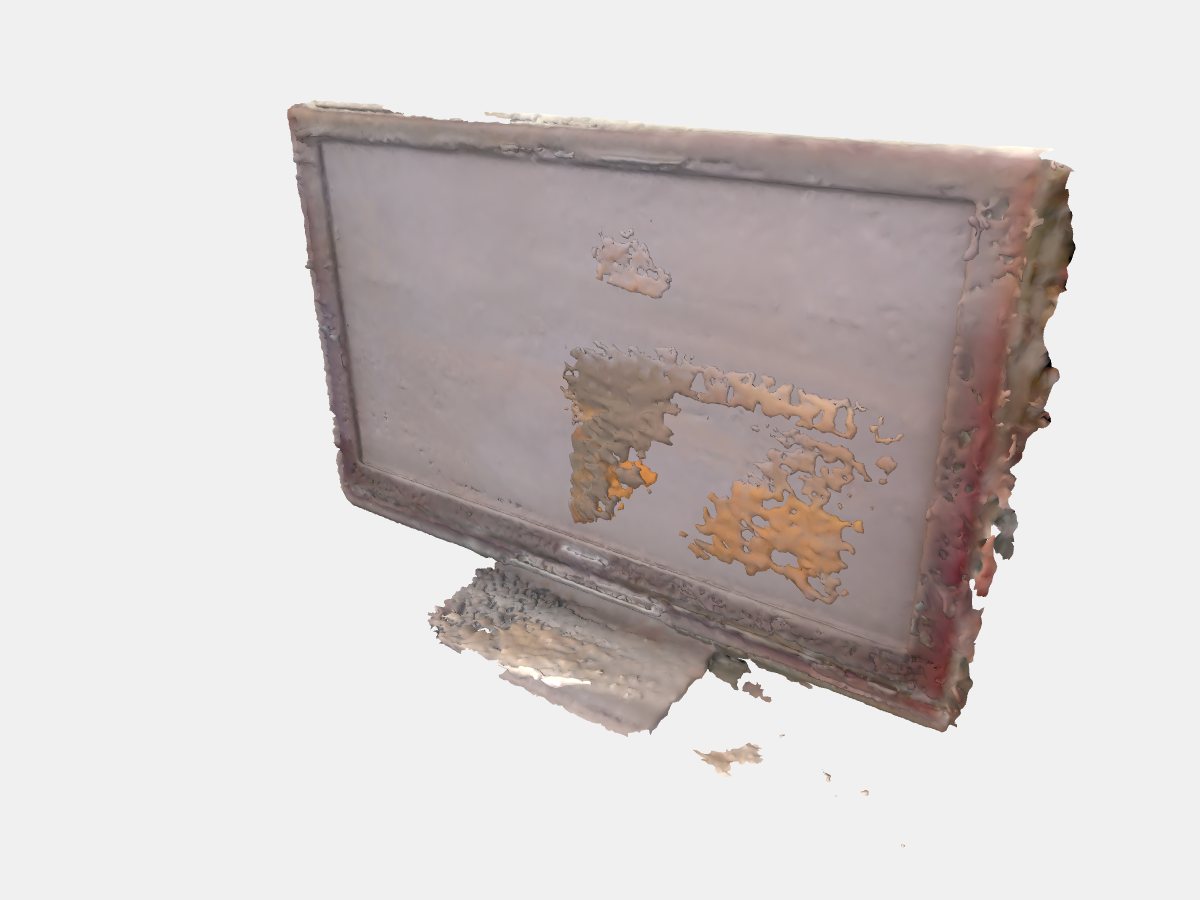}
\caption{TSDF mesh reconstructions from TrianguLang segmentations on ScanNet++ scenes. Left: ``sofa chair,'' Center: ``coffee table,'' Right: ``TV.'' Each mesh is extracted by fusing masked metric depth maps across 8 views using TSDF integration. The clean geometry demonstrates that TrianguLang produces view-consistent segmentations suitable for 3D reconstruction.}
\label{fig:tsdf_grid}
\end{figure}

Figure~\ref{fig:tsdf_grid} shows TSDF mesh reconstructions extracted from TrianguLang's segmented depth maps. Given a text query, the model produces per-view masks and metric depth estimates that can be fused into a coherent 3D mesh via TSDF integration. These reconstructions demonstrate that TrianguLang's segmentations are geometrically consistent across views, enabling downstream 3D applications without per-scene optimization.

\end{document}